\documentclass[10pt,journal,compsoc]{IEEEtran}
\ifCLASSOPTIONcompsoc
  \usepackage[nocompress]{cite}
\else
  \usepackage{cite}
\fi

\usepackage[section]{placeins}
\usepackage{cite}
\usepackage{amsmath,amssymb,amsfonts}
\usepackage{algorithmic}
\usepackage{graphicx}
\usepackage{textcomp}
\usepackage{makecell}
\usepackage{multirow}
\usepackage{hyperref}
\usepackage{booktabs}
\usepackage{caption}
\usepackage{subcaption}
\usepackage[ruled, vlined, linesnumbered]{algorithm2e}
\usepackage{color, xcolor}
\usepackage{bigstrut}
\emergencystretch=\maxdimen
\usepackage{bigstrut}  %new added usepackage
\ifCLASSINFOpdf
\else
\fi
\begin{document}
\title{MB-TaylorFormer V2: Improved Multi-branch Linear Transformer Expanded by Taylor Formula for Image Restoration}

\author{\IEEEauthorblockN{Zhi Jin, Yuwei Qiu, Kaihao Zhang, Hongdong Li, Wenhan Luo} 
\thanks{\hspace*{0.6em} This work was supported in part by: the National Natural Science Foundation of China (Grant 62071500, U24A20251); Shenzhen Science and Technology Program (Grant JCYJ20230807111107015).}
\thanks{%
  \hspace*{0.6em}
  Zhi Jin is School of Intelligent Systems Engineering, Shenzhen Campus of Sun Yat-sen University, Shenzhen, Guangdong 518107, P.R. China; with Guangdong Provincial Key Laboratory of Fire Science and Technology, Guangzhou 510006, P.R. China (e-mail: jinzh26@mail.sysu.edu.cn)
  
  Yuwei Qiu is with the School of Intelligent Systems Engineering, Shenzhen Campus of Sun Yat-sen University,  Shenzhen, 518107, Guangdong, P.R. China (e-mail: qiuyw9@mail2.sysu.edu.cn)

  Kaihao Zhang is with the School of Computer Science and Technology, Shenzhen Campus of Harbin Institute of Technology, Shenzhen, Guangdong 518055, P.R. China (super.khzhang@gmail.com).

  Hongdong Li is with the School of Engineering and Computer Science, Australian National University, Canberra, ACT 2600, Australia (hongdong.li@gmail.com).

   Wenhan Luo is with the Hong Kong University of Science and Technology, Clear Water Bay, Hong Kong (whluo.china@gmail.com).
}
\thanks{\hspace*{0.6em} Corresponding authors: Wenhan Luo, Kaihao Zhang.}
}
    


\markboth{Journal of \LaTeX\ Class Files,~Vol.~14, No.~8, August~2015}%
{Shell \MakeLowercase{\textit{et al.}}: Bare Demo of IEEEtran.cls for Computer Society Journals}



\IEEEtitleabstractindextext{%
\begin{abstract}
Recently, Transformer networks have demonstrated outstanding performance in the field of image restoration due to the global receptive field and adaptability to input. However, the quadratic computational complexity of Softmax-attention poses a significant limitation on its extensive application in image restoration tasks, particularly for high-resolution images. To tackle this challenge, we propose a novel variant of the Transformer. This variant leverages the Taylor expansion to approximate the Softmax-attention and utilizes the concept of norm-preserving mapping to approximate the remainder of the first-order Taylor expansion, resulting in a linear computational complexity. Moreover, we introduce a multi-branch architecture featuring multi-scale patch embedding into the proposed Transformer, which has four distinct advantages: 1) various sizes of the receptive field; 2) multi-level semantic information; 3) flexible shapes of the receptive field; 4) accelerated training and inference speed. Hence, the proposed model, named the second version of Taylor formula expansion-based Transformer (for short MB-TaylorFormer V2) has the capability to concurrently process coarse-to-fine features, capture long-distance pixel interactions with limited computational cost, and improve the approximation of the Taylor expansion remainder. Experimental results across diverse image restoration benchmarks demonstrate that MB-TaylorFormer V2 achieves state-of-the-art performance in multiple image restoration tasks, such as image dehazing, deraining, desnowing, motion deblurring, and denoising, with very little computational overhead. The source code is available at \href{https://github.com/FVL2020/MB-TaylorFormerV2}{\textcolor{blue}{https://github.com/FVL2020/MB-TaylorFormerV2}}.
\end{abstract}

\begin{IEEEkeywords}
Image restoration,  
linear Transformer,
Taylor formula,
multi-branch structure, 
multi-scale patch embedding
\end{IEEEkeywords}}

\maketitle

\IEEEdisplaynontitleabstractindextext

% \fi
\IEEEpeerreviewmaketitle

\IEEEraisesectionheading{\section{Introduction}\label{sec:introduction}}

\IEEEPARstart{T}{he} evolution of image restoration techniques has shifted from strategies reliant on prior information~\cite{he2010single} to deep learning-based models. Over the past decade, advancements in deep image restoration networks, characterized by sophisticated enhancements like multi-scale information fusion~\cite{ren2016single,qu2019enhanced}, refined convolution variants~\cite{wu2021contrastive}, and attention mechanisms~\cite{qin2020ffa}, have significantly improved performance. Recently, the Transformer architecture has been widely used in computer vision tasks~\cite{dosovitskiy2020image, chen2021pre}. However, there are two
challenges when applied in image restoration tasks: 1) the quadratic computational complexity of Transformer; 2)  the fixed-scale tokens generated by existing visual Transformer networks~\cite{zamir2022restormer,wang2022uformer} generally through fixed convolution kernels. Thus, further innovation is required to address these challenges.

\begin{figure}[t]
  \centering
  \captionsetup[subfloat]{labelformat=empty} % Adjust labelsep as needed
  \subfloat[(a) Dehazing]{
  \vspace{-0.2cm}
    \includegraphics[width=0.5\columnwidth]{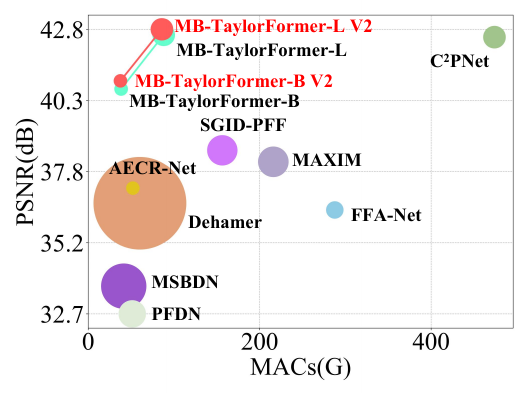}
  }
  \subfloat[(b) Deraining]{
  \vspace{-0.2cm}
    \includegraphics[width=0.5\columnwidth]{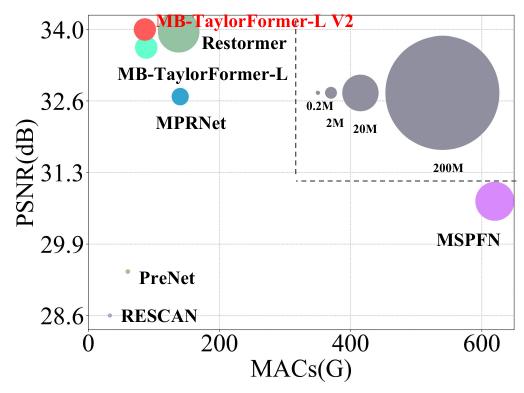}
  }\\
  \subfloat[(c) Deblurring]{
  \vspace{-0.2cm}
    \includegraphics[width=0.5\columnwidth]{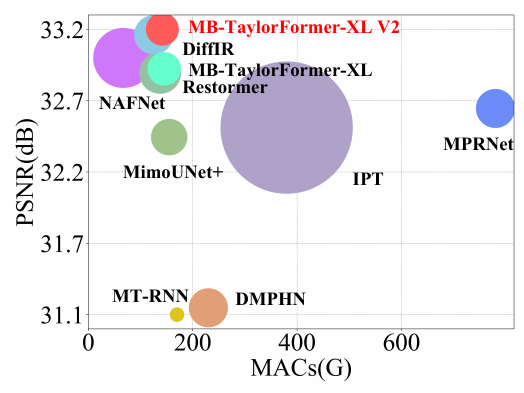}
  }
  \subfloat[(d) Denoising]{
  \vspace{-0.2cm}
    \includegraphics[width=0.5\columnwidth]{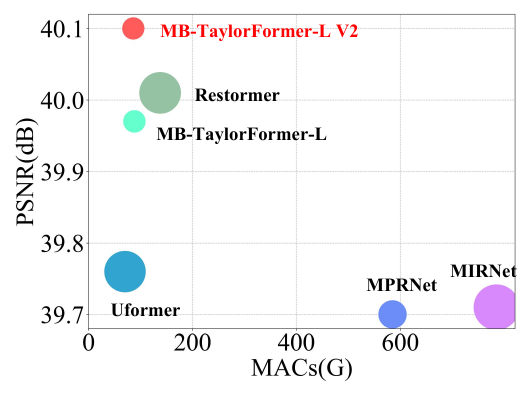}
  }
  \caption{\textbf{Improvement of MB-TaylorFormer V2 over the SOTA approaches.} The circle size is proportional to the number of model parameters. }
  \vspace{-0.5cm}
  \label{fig:comparison_four}
\end{figure}

For the first challenge, previous works have reduced the computational complexity of Transformer through various methods, such as shifted window~\cite{liu2021swin}, channel self-attention~\cite{zamir2022restormer}, and kernel functions~\cite{katharopoulos2020transformers}. However, these approaches often lead to some shortcomings, such as reduced receptive field, a lack of interaction between pixels, value approximation deficiencies, and attention focus challenges. Therefore, we propose a second version of Taylor formula expansion-based Transformer, named TaylorFormer V2. This variant applies a novel attention mechanism, termed Taylor Expanded Multi-Head Self-Attention++ (T-MSA++), on the entire feature map across spatial dimensions. %Importantly, it maintains linear computational complexity and approximates softmax attention both value and function.
Specifically, T-MSA++ comprises two components: the first term involves the first-order Taylor expansion of Softmax-attention, offering an approximation of its numerical values; the second term represents an approximation to the first-order remainder of the Taylor expansion, enabling the attention function of T-MSA++ to exhibit non-linearity and thus focusing more on crucial regions. Furthermore, we leverage the associative law of matrix multiplication to reduce the computational complexity of self-attention from $\mathcal{O}(n^{2})$ to $\mathcal{O}(n)$. This approach offers three distinct advantages: 1) it preserves the capacity of Transformer to model long-range dependencies in data; 2) it delivers accurate approximations of values and more focused attention; 3) it directs the self-attention towards pixel-level interactions rather than channel-level ones, enabling more nuanced processing of features.

For the second challenge, MPViT~\cite{lee2022mpvit} tackles the challenge by using multi-scale patches through parallel convolutional branches. However, we discover that its flexibility can be further improved. Taking inspiration from the success of DCN~\cite{dai2017deformable} and inception modules~\cite{szegedy2016rethinking} in CNN-based restoration networks, we introduce a multi-branch encoder-decoder backbone into the current TaylorFormer V2, and form MB-TaylorFormer V2, which is built on a multi-scale patch embedding. This embedding offers diverse receptive field sizes, multi-level semantic information, and flexible receptive field shapes. Furthermore, as the computational complexity of the Transformer is quadratic with channel dimension, the design of multi-branch allows for the use of fewer channels to further reduce computational cost. The multi-scale patch embedding generates tokens with varying scales and dimensions. The tokens from different scales are then simultaneously fed into different branches and finally fused. 

In summary, our primary contributions are as follows: (1) we use Taylor formula to perform a first-order Taylor expansion of Softmax-attention so that it satisfies the associative law of matrix multiplication, enabling the modeling of long-distance interactions between pixels with linear computational complexity; (2) based on the norm-preserving mapping, we approximate the higher-order terms of the Taylor expansion with linear computational complexity, which solves the unfocused attention problem of first-order expansion of Softmax-attention; (3) we devise a multi-branch architecture incorporating multi-scale patch embedding. This design, featuring multiple field sizes, a flexible shape of the receptive field, and multi-level semantic information, simultaneously processes tokens with different scales; (4) the experimental results of the image dehazing, deraining, desnowing, motion blurring and denoising tasks show that the proposed MB-TaylorFormer V2 achieves the state-of-the-art (SOTA) performance with less computational complexity and smaller number of parameters.

This work constitutes an extension of our conference paper published in ICCV 2023~\cite{qiu2023mb}. In comparison to our prior work, we have involved a significant amount of new content and additional experiments. (1) We deconstruct Softmax-attention using Taylor Expansion. Based on our research findings, we optimize the formula for T-MSA and redesign the network structure, introducing a more focused version referred to as T-MSA++. %This enhancement offers an improved version of the attention mechanism. 
(2) Given that T-MSA++ effectively addresses the limitations of T-MSA in approximating high-order remainder of the Taylor expansion, we remove the Multi-scale Attention Refine (MSAR) structure and adopt the convolutional position encoding to provide positional information and increase the rank of the attention map. (3) We implement parallel computations across multiple branches, enabling higher inference speeds on hardware. This achievement prompts researchers to consider accelerating their own multi-branch structures using parallel processing techniques. (4) As shown in Fig.~\ref{fig:comparison_four}, we validate the generalization capabilities of MB-TaylorFormer V2 on a broader range of image restoration tasks.

\section{Related Works}\label{sec:Related}

\begin{figure*}[htbp]
    \centering
    \includegraphics[width=17cm]{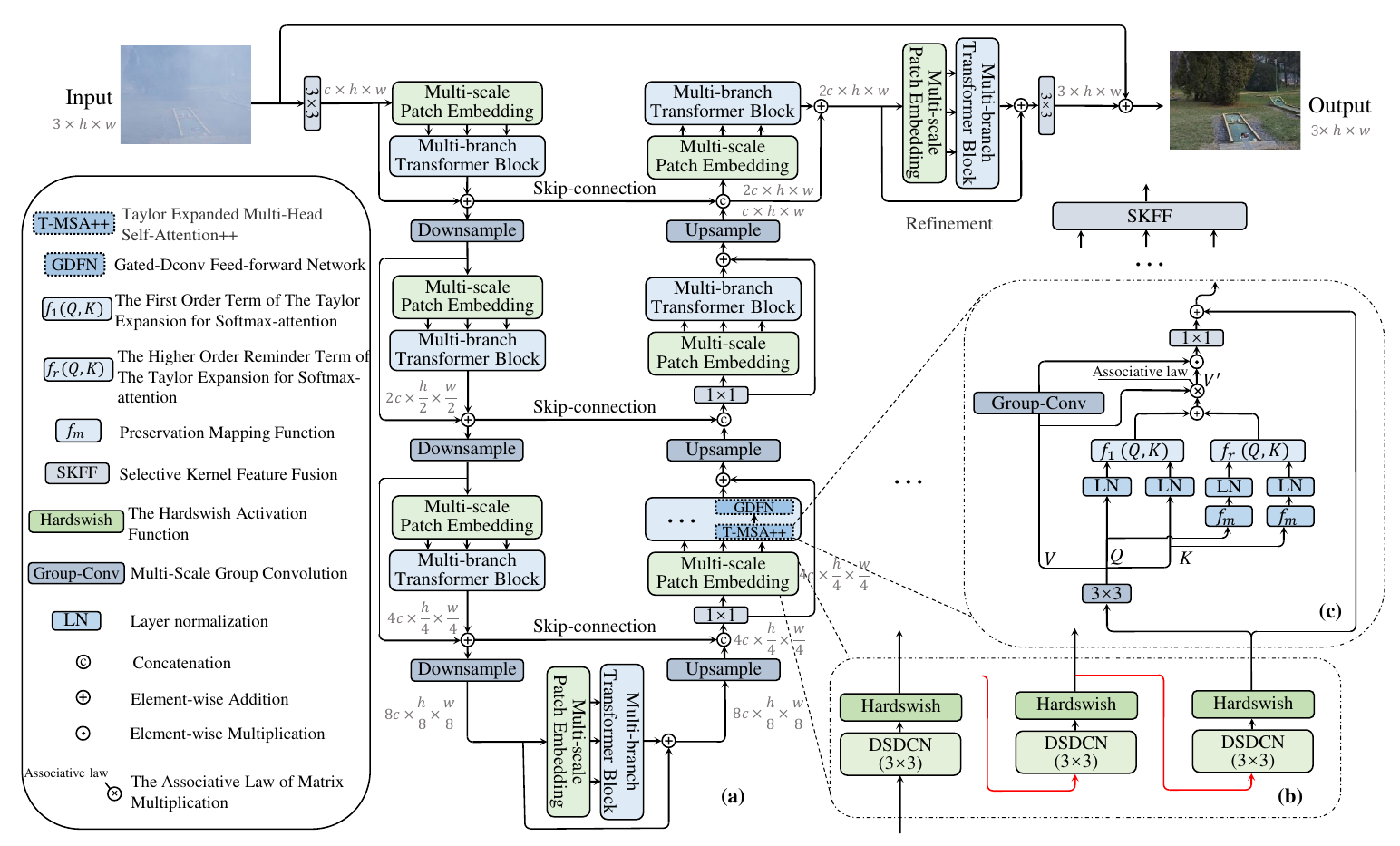}
    \caption{\textbf{Architecture of MB-TaylorFormer V2.} (a) MB-TaylorFormer V2 consists of the multi-branch hierarchical design based on multi-scale patch embedding. (b) Multi-scale patch embedding embeds coarse-to-fine patches. (c) T-MSA++ with linear computational complexity.} 
    \label{fig:pipeline}
\vspace{-0.2cm}
\end{figure*} 

\subsection{Image Restoration}
In recent research, there has been a notable shift towards employing data-driven CNN architectures~\cite{zhang2018density,jin2019flexible}  for image restoration, demonstrating their superior performance over traditional restoration methods~\cite{he2010single}. Among various CNN designs, considerable attention has been directed towards encoder-decoder-based U-Net architectures~\cite{cho2021rethinking} in the context of restoration. This preference is attributed to their hierarchical multi-scale representation, which proves effective in capturing intricate features while maintaining computational efficiency. Moreover, strategies involving attention mechanisms have emerged as a prominent avenue for image restoration, emphasizing the adaptive focus on different types of degraded regions~\cite{zhang2021dual}. The integration of generative adversarial networks (GANs) has also become increasingly popular, enabling the restoration of sharp image details with respect to the conventionally adopted metric of pixel-wise errors~\cite{zhang2018adversarial}. Some methods~\cite{zhang2021deep,dutta2024diva} based on physical priors have also garnered attention. For instance, Dutta et al.~\cite{dutta2024diva} introduce a deep neural network called DIVA, which unfolds a baseline adaptive denoising algorithm (De-QuIP). This approach leverages the theory of quantum many-body physics and achieves SOTA performance across various image restoration tasks. Additionally, as a novel and effective deep learning architecture, Transformer is receiving widespread attention from researchers in the field of image restoration. Yang et al.~\cite{yang2020learning} introduce the Transformer architecture in the task of image super-resolution, improving the detail information of the super-resolved images by reconstructing the self-attention relationships between high-resolution and low-resolution image texture details. Chen et al.~\cite{chen2021pre} propose a Transformer-based universal image restoration method, effectively enhancing the performance by utilizing pre-trained IPT model weights and fine-tuning on specific tasks. More improved versions of Transformer~\cite{zamir2022restormer,liang2021swinir,wang2022uformer} have been proposed. For a comprehensive overview of major design choices in image restoration, we recommend referring to reports from the NTIRE challenge~\cite{abuolaim2021ntire,ignatov2019ntire,nah2021ntire} and recent literature reviews~\cite{anwar2020deep,tian2020deep}.

\subsection{Efficient Self-attention}
The computational complexity of the Transformer increases quadratically with the growing spatial resolution of the feature map, placing a substantial demand on computational resources. Some approaches alleviate this burden by employing techniques, such as sliding window~\cite{hu2019local} or shifted window~\cite{wang2022uformer} based self-attention. However, these designs impose limitations on the ability of the Transformer to capture long-range dependencies in the data. MaxViT~\cite{tu2022maxvit} addresses the decrease in the receptive field with Grid attention, yet Grid attention still exhibits quadratic complexity on high-resolution images.

Another strategy involves modifying the attention mechanism of the vanilla Transformer. Restormer~\cite{zamir2022restormer} introduces self-attention between channels, but overlooks global interactions between pixels. Performer~\cite{choromanski2020rethinking} achieves linear complexity through the random projection, but the queries, keys, and values necessitate a large size, resulting in increased computational cost. Poly-nl~\cite{babiloni2021poly} establishes a connection between attention and high-order polynomials, yet this approach has not been explored in a self-attention structure. Other models~\cite{katharopoulos2020transformers,han2023flatten,qin2022cosformer} decompose the Softmax using kernel functions and leverage the associative law of matrix multiplication to achieve linear complexity. However, these models require constructing special kernel functions to approximate the functionality of Softmax-attention. e.g., ~\cite{katharopoulos2020transformers} demands that each element of the attention map is non-negative; ~\cite{qin2022cosformer} requires the attention map to exhibit local correlations; ~\cite{han2023flatten} necessitates more focused attention on relevant regions. Nonetheless, they all overlook numerical approximations.

\subsection{Multi-scale Transformer Networks}
In the field of high-level vision, in addition to simple pyramidal networks~\cite{zhang2020pyramid}, IFormer~\cite{si22022inception} integrates inception structures for blending high and low-frequency information. However, it neglects the utilization of varied patch sizes. CrossViT~\cite{chen2021crossvit} and MPViT~\cite{lee2022mpvit} handle multi-scale patches through multiple branches, aiming to achieve diverse receptive fields. Nonetheless, the flexibility of the receptive field shape is constrained due to fixed-shape convolutional kernels. In the domain of low-level vision, MSP-Former~\cite{yang2022mstfdn} employs multi-scale projections to assist Transformers in capturing complex degraded environments. Giqe~\cite{shyam2022giqe} employs a multi-branch approach to process feature maps of varying sizes. ~\cite{zhao2021complementary} employs multiple sub-networks to capture diverse features relevant to the task. GridFormer~\cite{wang2023gridformer} designs a grid structure using a residual
dense transformer block to capture multi-scale information. The recent Transformer networks designed for restoration tasks~\cite{zamir2022restormer,wang2022uformer} construct uncomplicated U-net architectures employing single-scale patches. Nevertheless, these endeavors scarcely delve into the exploration of multi-scale patches and multi-branch architectures. While~\cite{kulkarni2023aerial} utilizes deformable convolution in self-attention, it is noteworthy that the number of sampling points in the convolution kernel remains fixed. In contrast, our multi-scale deformable convolution not only boasts flexible sampling points but also offers multi-level semantic information.

\begin{figure}[t]
    \centering
    \includegraphics[width=7cm]{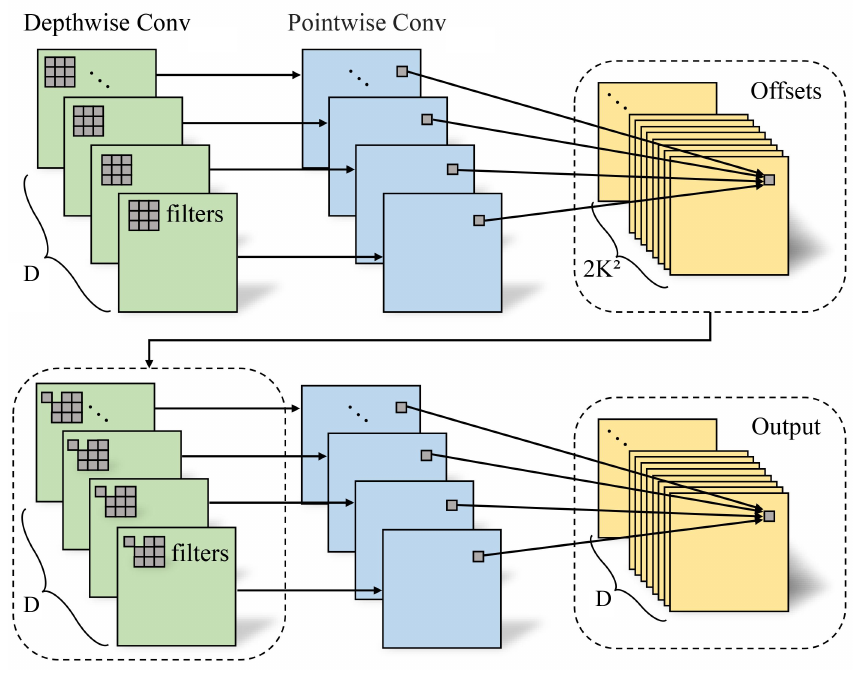}
    \caption{\textbf{Illustration of DSDCN.} The offsets are generated by $K \times K$ depthwise convolutions and pointwise convolutions, and the output is generated by $K \times K$ depthwise deformable convolutions and pointwise convolutions. D represents the number of channels of the feature maps.}
    \label{fig:figure3}
    \vspace{-0.2cm}
\end{figure} 
    \begin{figure}[t]
    \centering
    \includegraphics[width=8cm]{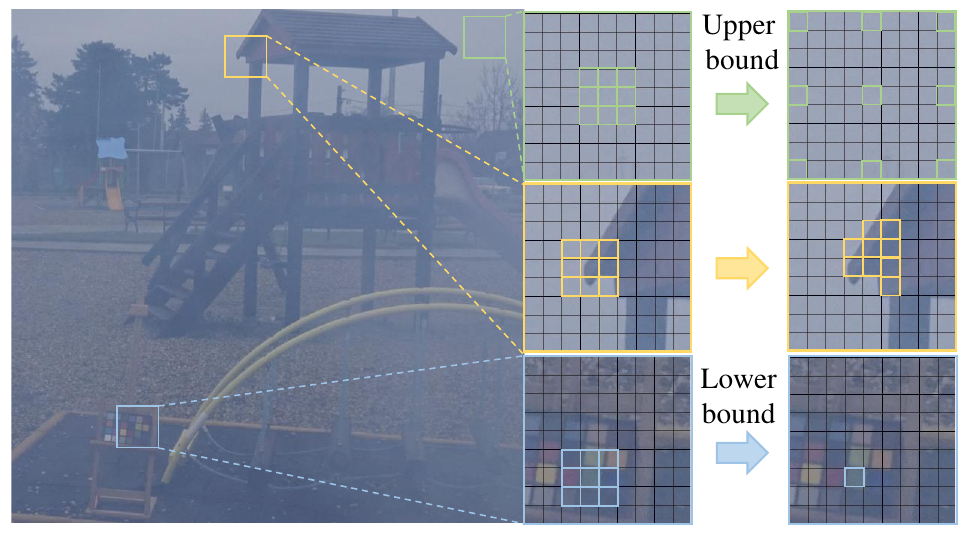}
    \caption{\textbf{Illustration of the receptive field of DSDCN (the offsets are truncated to [-3,3]).} The upper bound of the receptive field of the DSDCN is $9\times 9$ and the lower bound is $1\times 1$.}
    \label{fig:figure4}
    \vspace{-0.2cm}
\end{figure} 
\section{Method}
We aim to further explore the practicality of Transformer-based networks in image restoration tasks. To mitigate computational complexity, we employ Taylor expansion of Softmax-attention to use the associative law. Additionally, we present a technique for estimating the higher-order remainder of the Taylor expansion to focus on more crucial areas within the image. In the subsequent sections, we first present the architecture of MB-TaylorFormer V2 (Fig.~\ref{fig:pipeline}(a)). Then, we introduce multi-scale patch embedding (Fig.~\ref{fig:pipeline}(b)) and Taylor-expanded self-attention++ (Fig.~\ref{fig:pipeline}(c))

\subsection{Multi-branch Backbone}
Given a degraded image $ I\in \mathbb{R}^{3\times h\times w} $, we perform convolution for shallow feature extraction to generate $F_{\textbf{o}}\in \mathbb{R}^{c\times h\times w} $. Following this, a four-stage encoder-decoder network is employed for deep feature extraction. In each stage, a residual block is incorporated, comprising a multi-scale patch embedding and a multi-branch Transformer block, we replace the FFN layer in the Transformer with SKFF~\cite{zamir2022restormer}, which can adaptively select and fuse features from multiple kernel sizes, enabling the network to effectively capture multi-scale information. Utilizing multi-scale patch embedding, we generate tokens with various scales, which are then fed into multiple Transformer branches simultaneously. Each Transformer branch is composed of multiple Transformer encoders, and different branches perform parallel computations. At the end of the multi-branch Transformer block, we apply the SKFF module~\cite{zamir2020learning} to merge features generated by different branches, selectively fusing complementary features through an attention mechanism. Benefiting from this design, we can distribute the channel numbers across multiple branches. In general, the computational complexity of T-MSA++ increases quadratically with the growth of channel numbers, and the number of channels is much smaller than the number of tokens. Moreover, the divide-and-conquer approach of decomposing channels into multiple branches further reduces the overall computational cost. We utilize pixel-unshuffle and pixel-shuffle operations~\cite{shi2016real} in each stage for downsampling and upsampling features, respectively. Skip connections~\cite{ronneberger2015u} are employed to integrate information from the encoder and decoder, and a $ 1\times 1 $ convolutional layer is used for dimensionality reduction (except for the first stage). A residual block is also applied after the encoder-decoder structure to refine the fine structural and textural details. Finally, a $ 3\times 3$ convolutional layer is employed to reduce channel numbers and produce a residual image $ R\in \mathbb{R}^{3\times h\times w}$. The restored image is obtained as $I'= I+R$. To further reduce the computational cost, we incorporate depthwise separable convolutions~\cite{chollet2017xception} in the model.
%These design choices yielded improvements in quality, as we will see in the experimental section()%

\begin{figure}[t]
    \centering
    \includegraphics[width=4cm]{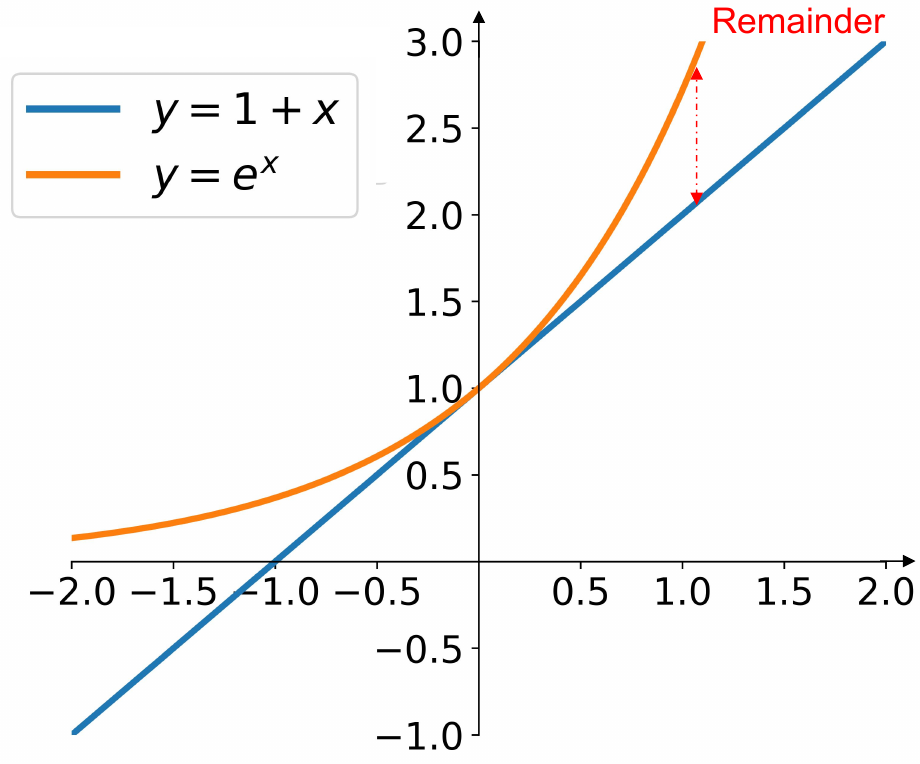}

    \caption{\textbf{$\mathbf{e^{x}}$ (orange) and its first-order Taylor expansion curve (blue).} The closer the value of $x$ to 0, the tighter the approximation of the orange line to the blue line.}
    \label{fig:figure5}
    \vspace{-0.4cm}
\end{figure} 

\begin{figure*}
  \centering
  \begin{subfigure}{0.60\textwidth}
    \includegraphics[width=\linewidth]{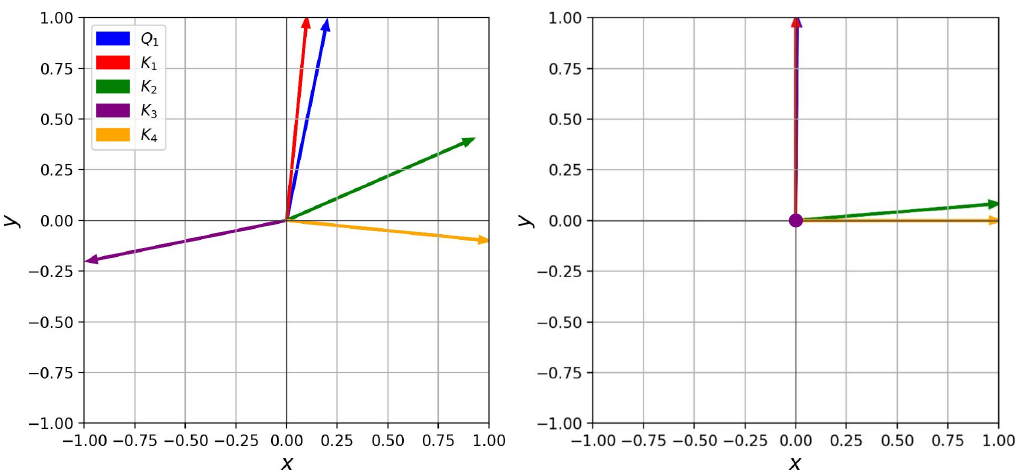}
    \caption{Query and keys before and after mapping}
    \label{fig:subfig_a}
  \end{subfigure}
  \quad
  \begin{subfigure}{0.34\textwidth}
    \includegraphics[width=\linewidth]{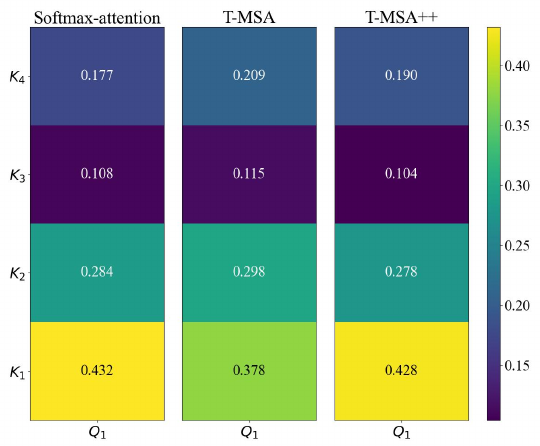}
    \caption{Attention maps}
    \label{fig:subfig_b}
  \end{subfigure}
  \caption{\textbf{Principles of our mapping function.} (a) Before mapping: Q=[0.2000, 0.9798], K1=[0.1000, 0.9950], K2=[0.9165, 0.4000], K3=[-0.9798, -0.2000], K4=[0.995, -0.1000]. After mapping: Q=[0.0083, 0.9999], K1=[0, 1], K2=[0.9966, 0.0828], K3=[0, 0], K4=[1, 0]. (b) The values of the attention map represent the dot product of Q and K.}
  \label{fig:main_figure}
\end{figure*}

\subsection{Multi-scale Patch Embedding}
Visual tokens exhibit considerable variation in scale. Previous approaches~\cite{wang2022uformer,liang2021swinir,zamir2022restormer} commonly use convolutions with fixed kernels for patch embedding, potentially resulting in a single scale of visual tokens. To tackle this limitation, we introduce a novel multi-scale patch embedding with three key properties: 1) various sizes of the receptive field; 2) flexible shapes of the receptive field; 3) multi-level semantic information.

Specifically, we employ multiple DCN \cite{dai2017deformable} layers with different scales of convolution kernels. This allows the patch embedding to generate visual tokens with varying coarseness and fineness, as well as facilitating flexible transformation modeling. Inspired by the concept of stacking conventional layers to expand receptive fields \cite{simonyan2014very}, we stack several DCN layers with small kernels instead of using a single DCN layer with large ones. This approach not only increases network depth and provides multi-level semantic information, but also aids in reducing number of parameters and computational load. All DCN layers are followed by Hardswish~\cite{howard2019searching} activation functions.

Similar to the approach used in depthwise separable convolutions~\cite{howard2017mobilenets, chollet2017xception}, we introduce a novel technique named depthwise separable and deformable convolutions (DSDCN). This method decomposes the components of DCN into depthwise convolution and pointwise convolution, as illustrated in Fig.~\ref{fig:figure3}. The computational costs for both standard DCN and DSDCN of an image with the resolution $h\times w$ are as follows:
\begin{equation}    
 \Omega(\mathrm{DCN})=2D K^{4} h w+D^{2} K^{2} h w +4D K^{2} h w,
\end{equation}
 \vspace{-0.5cm} 
 \begin{equation}
 \begin{split}
 \Omega(\mathrm{DSDCN})=8D K^{2} h w+D^{2} h w,
  \end{split}
\end{equation}
where $D$ is the number of channels of the feature maps, and $K$ denotes the kernel size. Compared to DCN, DSDCN significantly reduces computational complexity.

Given that images usually exhibit local relevance, and patch embedding captures the fundamental elements of feature maps, the visual elements (i.e., tokens) should be more focused on local areas. To control the receptive field range of the patch embedding layer, we truncate the offsets, which are pragmatically chosen to be within the range $[-3,3]$. As illustrated in Fig.~\ref{fig:figure4}, depending on the shape of the visual object, the model can autonomously select the receptive field size through learning. This selection process has an upper bound as $ 9\times 9 $, equivalent to a dilated convolution\cite{YuKoltun2016} with a dilation factor of $4$, and a lower bound as $ 1\times 1 $. In the case of setting up multi-scale patch embedding in parallel, the sizes of the receptive field for different branches are $ x_{1}\in [1,9] $, $ x_{2}\in [x_{1},x_{1}+8] $ and $ x_{3}\in [x_{2},x_{2}+8] $ in ascending order (for three branches). Experiments in Tab.~\ref{tab:tab-0} demonstrate that appropriately constraining the receptive field of each token can enhance the performance.

\subsection{Taylor Expanded Multi-head Self-Attention}
Let queries ($Q$), keys ($K$), and values ($V$) represent sequences of $h \times w$ feature vectors with dimensions $D$, respectively. The formula of the origin Transformer~\cite{vaswani2017attention} is as follows:
\begin{equation}
V' = \text{Softmax}\left(\frac{QK^{T}}{\sqrt{D}}\right)V.\label{con:att}
\end{equation}
Given that $Q \in \mathbb{R}^{hw \times D}$, $K \in \mathbb{R}^{hw \times D}$, and $V \in \mathbb{R}^{hw \times D}$, the application of Softmax results in a computational complexity for self-attention of $\mathcal{O}\left(h^{2}w^{2}\right)$, leading to high computational costs.

To reduce the computational complexity of self-attention from $\mathcal{O}\left(h^{2}w^{2}\right)$ to $\mathcal{O}\left(hw\right)$, we initially express the generalized attention equation for Eq.~\ref{con:att} as follows:
\begin{equation}    
V_{i}^{\prime}=\frac{\sum_{j=1}^{N} f\left(Q_{i}, K_{j}\right) V_{j}}{\sum_{j=1}^{N} f\left(Q_{i}, K_{j}\right)},\label{con:att2}
\end{equation}
where the matrix with $i$ and $j$ as subscript is the vector of the $i$-th and $j$-th row of matrix, respectively. $f(\cdot)$ denotes any mapping function. Eq.~\ref{con:att2} turns to Eq.~\ref{con:att} when we let $f\left(Q_{i}, K_{j}\right)=\exp \left(\frac{Q_{i} K_{j}^{T}}{\sqrt{D}}\right)$.
    If we apply the Taylor formula to perform a first-order Taylor expansion on $\exp \left(\frac{Q_{i} K_{j}^{T}}{\sqrt{D}}\right)$ at $0$, we can rewrite Eq.~\ref{con:att2} as:
\begin{equation}    
V_{i}^{\prime}=\frac{\sum_{j=1}^{N} \left(1+Q_{i}K_{j}^{T}+o\left(Q_{i}K_{j}^{T}\right) \right)V_{j}}{\sum_{j=1}^{N} \left(1+Q_{i}K_{j}^{T}+o\left(Q_{i}K_{j}^{T}\right)\right)}.\label{con:att3}
\end{equation}

To approximate $\exp \left(\frac{Q_{i} K_{j}^{T}}{\sqrt{D}}\right)$ and ensure that the weights in the attention map remain consistently greater than $0$, we normalize the magnitudes of $Q_{i}$ and $K_{j}$ to 1, generating $\tilde{Q_{i}}$ and $\tilde{K_{j}}$. We obtain the expression for the Taylor expansion of self-attention as follows:
\begin{equation}    
V_{i}^{\prime}=\frac{\sum_{j=1}^{N} \left(1+\tilde{Q}_{i}\tilde{K}^{T}_{j}+o\left(\tilde{Q}_{i}\tilde{K}^{T}_{j}\right) \right)V_{j}}{\sum_{j=1}^{N} \left(1+\tilde{Q}_{i}\tilde{K}^{T}_{j}+o\left(\tilde{Q}_{i}\tilde{K}^{T}_{j}\right)\right)}.\label{con:att3}
\end{equation}
If we neglect the higher-order terms of the Taylor expansion, we can simplify Eq.~\ref{con:att3} and leverage the associativity of matrix multiplication to reduce computational complexity, as shown below:
\begin{equation}   
\begin{split}
V_{i}^{\prime}=\frac{\sum_{j=1}^{N} V_{j}+\tilde{Q_{i}}\sum_{j=1}^{N} \tilde{K}_{j}^{T}V_{j}}{N+\tilde{Q}_{i}\sum_{j=1}^{N} \tilde{K}_{j}^{T}}.\label{con:att5}
\end{split}
\end{equation}

However, ignoring the higher-order terms in the Taylor expansion of Softmax-attention typically sacrifices the non-linear characteristics of the attention map, reducing the attention capability of the model to some important regions in the image. In the next section, we introduce how we predict the remainder of Softmax-attention, ensuring that the attention map of T-MSA++ retains non-linear characteristics while maintaining linear computational complexity.
%At this point, we have completed the Taylor expansion of softmax-attention. And we can observe that the computational complexity of the first two terms in the numerator and denominator of Eq.~\eqref{con:att4} is $\mathcal{O}\left(hw\right)$, while the computational complexity of the remainder term is $\mathcal{O}\left(h^{2}w^{2}\right)$. The next crucial step is to approximate the remainder of the Taylor expansion with linear computational complexity.
%

\subsection{Focused Taylor Expansion Remainder}

From the analysis of Fig.~\ref{fig:figure5}, it can be concluded that the remainder $o\left(\tilde{Q}_{i}\tilde{K}_{j}^{T}\right)$ possesses two properties: 1) non-negativity; 2) offering a non-linear scaling of $\tilde{Q}_{i}\tilde{K}_{j}^{T}$ to provide more focused attention. Therefore, we have established a new mapping function as follows:
\begin{equation}   
\begin{split}
o\left(\tilde{Q}_{i}\tilde{K}_{j}^{T}\right)&=\phi_{p}(\tilde{Q}_{i})\phi_{p}^{T}(\tilde{K}_{j}),\\
where\quad   \phi_{p}(x)&=\frac{\left \| \text{ReLU}(x) \right \| }{\left \|\text{ReLU}(x^{p})\right \| } \text{ReLU}(x^{p}),
\end{split}\label{con:att5}
\end{equation}
where $x^{p}$ represents the element-wise power $p$ of $x$. We adopt the ReLU function similar to previous linear attention modules to ensure the non-negativity of the input and the validity of the denominator in Eq.~\ref{con:att5}. A direct observation reveals that the norm of the feature is maintained after the mapping, i.e., $\left \| x \right \| = \left \| \phi_{p}( x) \right \|$, indicating that only the feature direction is adjusted.

Fig.~\ref{fig:main_figure}(a) presents the principles of our mapping function. It diminishes the cosine distance between $Q_{i}$ and $K_{j}$ when they have a small initial distance, and conversely, increases the cosine distance between them when the initial distance is large. The distinctive properties of this mapping function enable T-MSA++ to assign more significant weights to $Q_{i}$ and $K_{j}$ vectors with the increased similarity in the attention map. Consequently, T-MSA++ achieves a closer approximation to Softmax-attention, as depicted in Fig.~\ref{fig:main_figure}(b).

%Figure ~\ref{fig:subfig_a} introduces our mapping function, which, when q and k have a smaller cosine distance, can further reduce the cosine distance between q and k. Conversely, the mapping function can increase the cosine distance between q and k. Thanks to the properties of this mapping function, T-MSA++ enables the attention map to assign greater weights to Qi and Kj vectors with higher similarity, making it closer to Softmax-attention in terms of approximation, as illustrated in Figure~\ref{fig:subfig_b}.

Obviously, through the mapping function, $o\left(\tilde{Q}_{i}\tilde{K}_{j}^{T}\right)$ satisfies the following properties: 1) non-negativity; 2) when $p>1$, for larger values of $\tilde{Q}_{i}\tilde{K}^{T}_{j}$, the following relationship exists:
\begin{equation}   
\begin{split}
 \left \| \phi_{p}( Q_{i})\phi_{p}^{T}( K_{j}) \right \|> \left \|Q_{i} K_{j}^{T}\right \|,
\end{split}
\end{equation}
for smaller values of $\tilde{Q}_{i}\tilde{K}^{T}_{j}$, the following relationship holds:
\begin{equation}   
\begin{split}
 \left \| \phi_{p}( Q_{i})\phi_{p}^{T}( K_{j}) \right \|<  \left \|Q_{i} K_{j}^{T}\right \|.
\end{split}
\end{equation}

\IncMargin{1em}

To prevent the focused Taylor expansion reminder from excessively disrupting the numerical approximation of T-MSA++ to Softmax-attention, we introduce a learnable modulation factor 's' before $\phi_{p}(\tilde{Q}_{i})\phi_{p}^{T}(\tilde{K}_{j})$, which is initialized to 0.5 and can be learned during the model training process. Furthermore, the following formula can be derived from Eq.~\ref{con:att3} as
\begin{equation}   
\begin{split}
V_{i}^{\prime}&=\sum_{j=1}^{N}f_{1}(Q_{i},K_{j})V_{j}+\sum_{j=1}^{N}f_{r}(Q_{i},K_{j})V_{j}
\\&=\frac{\sum_{j=1}^{N} V_{j}+\tilde{Q_{i}}\sum_{j=1}^{N} \tilde{K}_{j}^{T}V_{j}}{N+\tilde{Q}_{i}\sum_{j=1}^{N} \tilde{K}_{j}^{T}+s\cdot \phi_{p}(\tilde{Q}_{i})\sum_{j=1}^{N}\phi_{p}^{T}(\tilde{K}_{j})}\\&+\frac{s\cdot \phi_{p}(\tilde{Q}_{i})\sum_{j=1}^{N}\phi_{p}^{T}(\tilde{K}_{j})V_{j}}{N+\tilde{Q}_{i}\sum_{j=1}^{N} \tilde{K}_{j}^{T}+s\cdot \phi_{p}(\tilde{Q}_{i})\sum_{j=1}^{N}\phi_{p}^{T}(\tilde{K}_{j})}
\end{split}
\end{equation}

\begin{algorithm}
  \SetKwData{Left}{left}\SetKwData{This}{this}\SetKwData{Up}{up}
  \SetKwFunction{Union}{Union}\SetKwFunction{FindCompress}{FindCompress}
  \SetKwInOut{Input}{input}\SetKwInOut{Output}{output}

\noindent{\Input{A feature map $I_{f}$ of shape $b\times h\times w \times D$}}
\noindent{\Output{A feature map $O_{f}$ of shape $b\times h\times w \times D$}}
  \BlankLine
\textcolor{green!70!black}{\# $Q, K, V: b\times head \times hw \times \frac{c}{head}$}

$Q, K, V=\text{rearrange}( \text{project}(I_{f}))$ \ 

$Q_{1} = \text{normalize}(Q, dim=-1)$

$K_{1} = \text{normalize}(K, dim=-1)$

$Q_{h} = \text{normalize}(\text{ReLU}(Q)**factor, dim=-1)$

$K_{h} = \text{normalize}(\text{ReLU}(K)**factor, dim=-1)$

    \textcolor{green!70!black}{\# mm: matrix multiplication}

    \textcolor{green!70!black}{\# $K^{T}: b\times head \times\frac{c}{head} \times hw$}
    \BlankLine
    $Q\_K\_V_{1}=\text{mm}(Q_{1}, \text{mm}(K^{T}_{1}, V))$

    $Q\_K\_V_{h}=\text{mm}(Q_{h}, \text{mm}(K^{T}_{h}, V))$
    
\textcolor{green!70!black}{\# $Ones$ represents a matrix with all values equal to 1}

    $Ones\_V_{h} = \text{sum}(V, dim=-2)\text{.unsqueeze}(2)$

    $K\_Ones_{1} = \text{sum}(K_{1}^{T}, dim=-2))\text{.unsqueeze}(2)$
        
    $Q\_K\_Ones_{1}=\text{mm}(Q, K\_Ones_{1})$

    $K\_Ones_{h} = \text{sum}(K_{h}^{T}, dim=-2))\text{.unsqueeze}(2)$
        
    $Q\_K\_Ones_{h}=\text{mm}(Q, K\_Ones_{h})$
    
    \textcolor{green!70!black}{\# $D, N: b\times head \times hw \times\frac{c}{head}$}

        $N=Ones\_V_{h} + Q\_K\_V_{1}+Q\_K\_V_{h}$
        
    $D=h\times w+ Q \_ K \_ Ones_{1} + Q \_ K \_ Ones_{h} +1e^{-6}$

 \textcolor{green!70!black}{\# $O': b \times h \times w \times c$}
    
    $O'$=rearrange(div($N$, $D$)) + CPE(V)

    $O_{f}$ = project($O'$)

  \caption{Pseudo code of T-MSA++ in a PyTorch-like style.}\label{algo_disjdecomp}

\end{algorithm}\DecMargin{1em}
\subsection{Convolutional Positional Encoding}

The self-attention mechanism is agnostic to positions, although some positional encoding methods~\cite{liu2021swin, vaswani2017attention} address this issue by incorporating positional information. However, these methods often require fixed windows or inputs. In T-MSA++, we employ a straightforward method called convolutional positional encoding (CPE). This method is a form of relative positional encoding that can be applied to input images of arbitrary resolutions. Specifically, for the input $V$, we utilize depthwise convolution (DWC) with multi-scale convolution kernels for performing grouped convolution, as shown below
\begin{equation}  
  \begin{split}
    V_{\mathrm{I}}, V_{\mathrm{II}}, \ldots = \text{Split}(V),
  \end{split}
\end{equation}  
\begin{equation}  
  \begin{split}
\text{CPE}(V) = \text{Cat}(\text{DWC}_{3\times 3}(V_{\mathrm{I}}), \text{DWC}_{5\times 5}(V_{\mathrm{II}}), \ldots).
\end{split}
\end{equation}  
We add it to the output $V{}'$ obtained from the previous section. The last formula is as follows:
\begin{equation}  
  \begin{split}
\text{T-MSA++}(Q, K, V)= V{}'+ \text{CPE}(V).
\end{split}
\end{equation} 

The computational costs for both Softmax-attention and T-MSA++ in the context of an input feature with resolution $h \times w$ are as follows:
\begin{equation}    
 \Omega(\text{Softmax-attention})=2 (h w)^{2} D + 4 h w D^{2},
\end{equation}
 \vspace{-0.5cm} 
 \begin{equation}
 \begin{split}
 \Omega(\text{T-MSA++})=8 h w D^{2} + 4 K^{2} h w D,
  \end{split}
\end{equation}
where $D$ is the number of input channels.

Algorithm~\ref{algo_disjdecomp} shows the pseudo-code for the matrix implementation of T-MSA++, which implements the efficient self-attention operations.

\textbf{Rank of the attention map matrix.} For general kernel models~\cite{katharopoulos2020transformers}, there is a constraint on the rank of their attention maps given by the following formula:
\begin{equation}  
  \begin{split}
\text{Rank}(\phi(Q)\phi^{T}(K)) &\le \min(\text{Rank}(\phi(Q)), \text{Rank}(\phi^{T}(K)))
\\& \le \min(hw, D),
\end{split}
\end{equation} 
where $D$ is usually much smaller than $hw$, especially for image restoration, so it is challenging to achieve full rank.

On the contrary, the attention map of T-MSA++ is more likely to achieve full rank. For better illustration, we ignore the normalization effect of the denominator in T-MSA++. Since denominator normalization involves proportional scaling of all elements in each row of the attention map, it does not affect the rank of the attention map. The simplified formula for calculating the attention map of T-MSA++ is as follows
\begin{equation}  
  \begin{split}
    M_{att}= 1+QK^{T} + \phi_{p}(Q)\phi_{p}^{T}(K)+ M_{DWC},
\end{split}
\end{equation} 
where $M_{att}$ and $M_{DWC}$ represent the simplified attention map of T-MSA++ and the sparse matrix corresponding to the DWC, respectively. Consequently, we can derive the following relationship
\begin{equation}  
  \begin{split}
\text{Rank}(M_{att}) &\le \text{min}(1 + \min(\text{Rank}(Q), \text{Rank}(K^{T})) \\&+ \min(\text{Rank}(\phi(Q)), \text{Rank}(\phi^{T}(K))) \\& + \text{Rank}(M_{DWC}), hw).
\end{split}
\end{equation}
In theory, achieving full rank through the learning of parameters $M_{DWC}$ is possible, enabling the attention map of T-MSA++ to have a higher rank. Consequently, in most cases, T-MSA++ exhibits richer feature representation.

\section{Experiments}
\subsection{Experiment Setup}

We assess the effectiveness of the proposed MB-TaylorFormer V2 across benchmark datasets for five distinct image restoration tasks: (a) image dehazing (b) image deraining, (c) image desnowing, (d) image motion deblurring, and (f) image denoising. 

\textbf{Implementation Details.} We present three variants of MB-TaylorFormer V2, namely MB-TaylorFormer-B V2 (the foundational model), MB-TaylorFormer-L V2 (a large variant), and MB-TaylorFormer-XL V2 (an extra large variant), the detailed structure is shown in Tab.~\ref{tab:jiegou}. Data augmentation is performed through random cropping and flipping. The initial learning rate is set to 3e-4 and is systematically decreased to 1e-6 using cosine annealing~\cite{loshchilov2016sgdr}. The loss functions include L1 loss and FFT loss~\cite{liu2020trident}. All compared methods are trained on the same training datasets and evaluated on the same testing datasets.

\renewcommand{\arraystretch}{1.3}
\begin{table*}[t]
 \centering
\caption{Detailed structural specification of three variants of our MB-TaylorFormer V2.}
\label{tab:jiegou}
\scalebox{1}{
\begin{tabular}{c|c|c|c|c|c}
\toprule
Model             & Num. of Branches      & Num. of Blocks        & Num. of Channels               & Params & MACs   \\ \hline
MB-TaylorFormer-B V2 & {[}2, 2, 2, 2, 2, 2, 2, 2{]} & {[}2, 3, 3, 4, 3, 3, 2, 2{]} & {[}24, 48, 72, 96, 72, 48, 24, 24{]}  & 2.63M  & 37.7G \\
MB-TaylorFormer-L V2 & {[}2, 3, 3, 3, 3, 3, 2, 2{]} & {[}4, 6, 6, 8, 6, 6, 4, 4{]} & {[}24, 48, 72, 96, 72, 48, 24, 24{]} & 7.29M   &86.0G \\
MB-TaylorFormer-XL V2 & {[}2, 3, 3, 3, 3, 3, 2, 2{]} & {[}4, 6, 6, 8, 6, 6, 4, 4{]} & {[}28, 56, 112, 160, 112, 56, 28, 28{]} & 16.26M   &141.9G \\
\bottomrule
\end{tabular}}
\end{table*}

\begin{table*}[t]
\centering
\caption{\textbf{Quantitative comparisons of benchmark models on dehazing datasets.} ``-" indicates that the result is not available. The best and second best results are highlighted in bold and underlined, respectively.}
    \label{tab:dehze}
\begin{tabular}{c|cc|cc|cc|cc|cc|cc}
\toprule
\multirow{2}{*}{Models} & \multicolumn{2}{c|}{SOTS-Indoor}& \multicolumn{2}{c|}{SOTS-Outdoor} & \multicolumn{2}{c|}{O-HAZE}     & \multicolumn{2}{c|}{Dense-Haze} & \multicolumn{2}{c|}{NH-HAZE}    & \multicolumn{2}{c}{Overhead}          \\ \cline{2-13} 
                & PSNR$\uparrow$            & SSIM$\uparrow$            & PSNR$\uparrow$            & SSIM$\uparrow$           & PSNR$\uparrow$           & SSIM$\uparrow$           & PSNR$\uparrow$                & SSIM$\uparrow$      & PSNR$\uparrow$           & SSIM$\uparrow$           & \multicolumn{1}{c|}{Params$\downarrow$} & MACs$\downarrow$   \\ \hline
PFDN~\cite{dong2020physics }    & 32.68           & 0.976      & -           & -                  & -              & -              & -                   & -         & -              & -              & \multicolumn{1}{c|}{11.27M}  & 51.5G  \\
MSBDN~\cite{dong2020multi }                      & 33.67           & 0.985    & 33.48          & 0.982       & 24.36          & 0.749          & 15.13               & 0.555     & 17.97          & 0.659          & \multicolumn{1}{c|}{31.35M}  & 41.5G \\
FFA-Net~\cite{qin2020ffa}                   & 36.39           & 0.989      & 33.57          & 0.984       & 22.12          & 0.770          & 15.70               & 0.549     & 18.13          & 0.647          & \multicolumn{1}{c|}{4.46M}   & 287.8G \\
AECR-Net~\cite{wu2021contrastive}                   & 37.17           & 0.990      & -              & -      & -              & -              & 15.80               & 0.466     & -              & -              & \multicolumn{1}{c|}{\textbf{2.61M}}   & 52.2G  \\
MAXIM-2S~\cite{tu2022maxim}                 & 38.11           & 0.991         & 34.19           & 0.985       & -              & -              & -                   & -         & -              & -              & \multicolumn{1}{c|}{14.10M}  & 216.0G \\
HCD~\cite{wang2024restoring}                     & 38.31           & 0.991     & -              & -          & -              & -              & 16.41               & 0.563     & -              & -              & \multicolumn{1}{c|}{5.58M}   & 104.0G \\
SGID-PFF~\cite{bai2022self}              & 38.52           & 0.991          & 30.20          & 0.975    & 20.96          & 0.741          & 12.49               & 0.517     & -              & -              & \multicolumn{1}{c|}{13.87M}  & 156.4G \\
Dehamer~\cite{guo2022image}                  & 36.63           & 0.988      & 35.18           & 0.986      & 25.11          & 0.777          & 16.62               & 0.560     & 20.66          & 0.684          & \multicolumn{1}{c|}{132.50M} & 60.3G  \\
$\mathrm{C^{2}PNet}$~\cite{zheng2023curricular}    & 42.56           & 0.995        & 36.68           & 0.990      & 25.20          & 0.785          & 16.88               & 0.573     & 20.24          & 0.687          & \multicolumn{1}{c|}{7.17M}   & 461.0G  \\
ConvIR-S~\cite{cui2024revitalizing}    & 41.53           & \underline{0.994}        & 37.95           &    0.990   & 25.25          &   0.784     & \textbf{17.45}              &    \underline{0.608}  & 20.65         &     0.692     & \multicolumn{1}{c|}{5.53M}   & 42.1G \\
ConvIR-B~\cite{cui2024revitalizing}    & \underline{42.72}           & \textbf{0.995}        & \textbf{39.42}           &   \textbf{0.992}    & \underline{25.36}         &     0.780     & 16.86               &   0.600   & 20.66          &     0.691      & \multicolumn{1}{c|}{8.63M}   & 71.2G \\\hline
Ours-B V1~\cite{qiu2023mb}        & 40.71           & 0.992    & 37.42           & 0.989       & 25.05          & 0.788          & 16.66               & 0.560     & 20.43          & 0.688          & \multicolumn{1}{c|}{2.68M}   & \underline{38.5G}  \\
Ours-L V1~\cite{qiu2023mb}        & 42.64     & \underline{0.994}    & 38.09     & \underline{0.991}    & 25.31    & 0.782          & 16.44               & 0.566     & 20.49          & 0.692          & \multicolumn{1}{c|}{7.43M}   & 88.1G  \\
Ours-B V2    & 41.00           & 0.993    & 37.81          & \underline{0.991}       & 25.29          & \underline{0.790}    & \underline{16.95}      & \textbf{0.621}     & \underline{20.73}    & \underline{0.703}    & \multicolumn{1}{c|}{\underline{2.63M}}   & \textbf{37.7G}  \\
Ours-L V2    & \textbf{42.84}  & \textbf{0.995}  & \underline{39.25}  & \textbf{0.992} & \textbf{25.43} & \textbf{0.792} & 16.90         & 0.607     & \textbf{20.77} & \textbf{0.705} & \multicolumn{1}{c|}{7.29M}   & 86.0G  \\ \bottomrule
\end{tabular}
\end{table*}

\begin{figure*}[t]
    \centering
    \includegraphics[width=18cm]{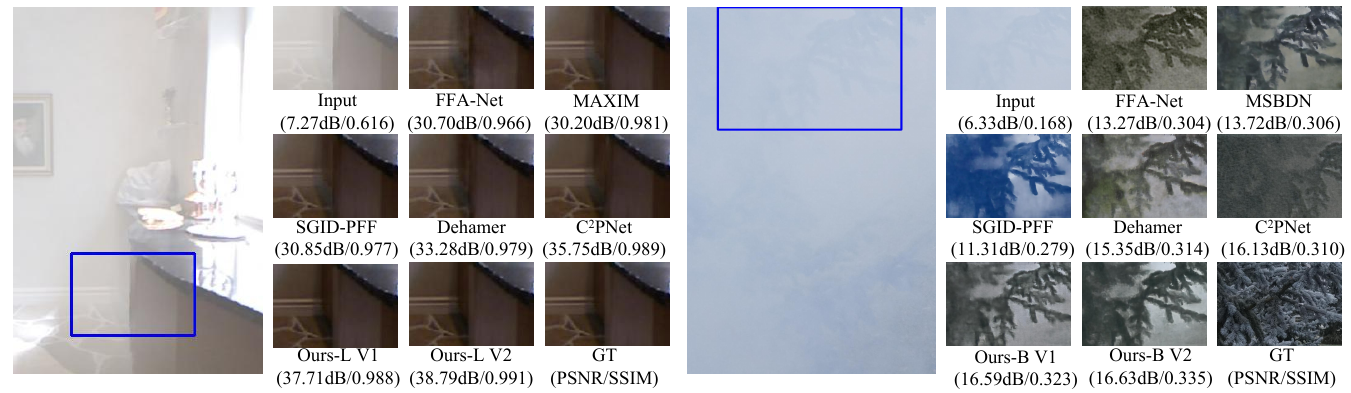}
  \caption{\textbf{Image dehazing on the images "00410" and "52\_hazy" from SOTS~\cite{8451944} and Dense-Haze~\cite{ancuti2019dense}.} Our MB-TaylorFormer-L V2 generates dehazed images with color fidelity and finer textures. ``Ours V1" represents the conference version of our MB-TaylorFormer~\cite{qiu2023mb}.}
    \label{fig:haze}
    \vspace{-0.3cm}
\end{figure*} 

\begin{table}[]
\centering
\caption{\textbf{Quantitative comparisons of benchmark models on HAZE4K datasets.} ``-'' indicates that the result is not available. The best and second best results are highlighted in bold and underlined, respectively.}
    \label{tab:dehze2}
\begin{tabular}{c|cc|cc}
\toprule
\multirow{2}{*}{Models} & \multicolumn{2}{c|}{Haze4K} & \multicolumn{2}{c}{Overhead} \\ \cline{2-5} 
                        & PSNR$\uparrow$         & SSIM$\uparrow$        & Params$\downarrow$       & MACs$\downarrow$         \\ \hline
                        MSBDN~\cite{dong2020multi}                & 22.99       & 0.85        & 31.35M            & \textbf{41.5G}            \\
                        FFA-Net~\cite{qin2020ffa}                & 26.96        & 0.95        & \textbf{4.46M}            & 287.8G            \\
DMT-Net~\cite{duong2023dmt}                 & 28.53        & 0.96         & -             & -            \\
PMNet~\cite{ye2022perceiving}                   & 33.49        & \underline{0.98}         & 18.90M        & 81.1G        \\
FSNet~\cite{cui2023image}                   & 34.12        & \textbf{0.99}         & 13.28M         & 110.5G        \\
ConvIR-S~\cite{cui2024revitalizing}                & 33.36        & \textbf{0.99}         & \underline{5.53M}          & \underline{42.1G}         \\
ConvIR-B~\cite{cui2024revitalizing}              & 34.15        & \textbf{0.99}         & 8.63M         & 71.2G        \\
ConvIR-L~\cite{cui2024revitalizing}                & \underline{34.50}        & \textbf{0.99}         & 14.83M         & 129.9G       \\ \hline
Ours-L V1               & 34.47        & \textbf{0.99}        & 7.43M          & 88.1G         \\
Ours-L V2               & \textbf{34.92}         & \textbf{0.99}       & 7.29M          & 86.0G        \\ \bottomrule
\end{tabular}
\end{table}

\begin{figure*}[t]
    \centering
    \includegraphics[width=18cm]{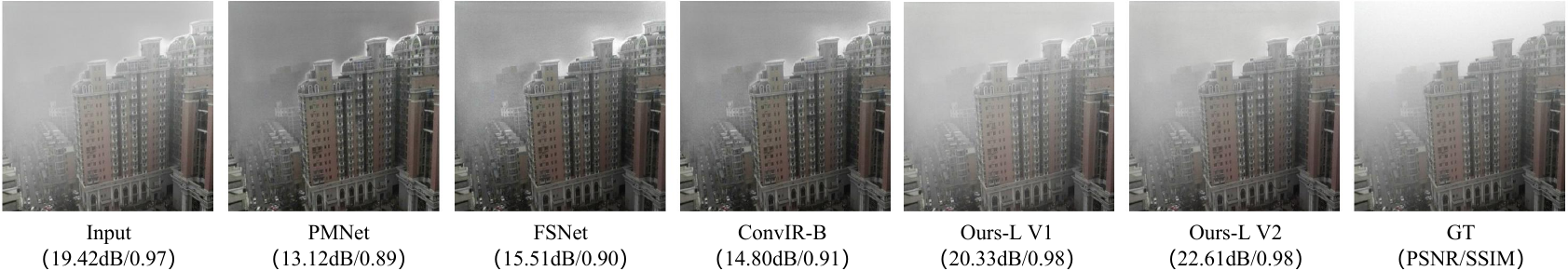}
  \caption{\textbf{Image dehazing on the images "443\_0.56\_1.04" from Haze4K~\cite{liu2021synthetic}. Our MB-TaylorFormer-L V2 generates dehazed images with fewer artifacts. }}
    \label{fig:dehaze2}
    \vspace{-0.3cm}
\end{figure*}

\begin{table*}[t]
\caption{\textbf{Quantitative comparisons of benchmark models on deraining datasets.} ``-" indicates that the result is not available. The best and second-best results are highlighted in bold and underlined, respectively.}
    \label{tab:derain}
    \centering
    \scalebox{0.95}{
\begin{tabular}{c|cc|cc|cc|cc|cc|cc}
\toprule
\multirow{2}{*}{Models} & \multicolumn{2}{c|}{Test100~\cite{zhang2019image}}    & \multicolumn{2}{c|}{Rain100H~\cite{yang2017deep}}   & \multicolumn{2}{c|}{Rain100L~\cite{yang2017deep}}   & \multicolumn{2}{c|}{Test2800~\cite{fu2017removing}}   & \multicolumn{2}{c|}{Test1200~\cite{zhang2018density}}   & \multicolumn{2}{c}{Average}     \\ \cline{2-13} 
                         & PSNR$\uparrow$ & SSIM$\uparrow$ & PSNR$\uparrow$ & SSIM$\uparrow$ & PSNR$\uparrow$ & SSIM$\uparrow$ & PSNR$\uparrow$ & SSIM$\uparrow$ & PSNR$\uparrow$ & SSIM$\uparrow$ & PSNR$\uparrow$ & SSIM$\uparrow$ \\ \hline
DerainNet~\cite{fu2017clearing}                & 22.77          & 0.810          & 14.92          & 0.592          & 27.03          & 0.884          & 24.31          & 0.861          & 23.38          & 0.835          & 22.48          & 0.796          \\
SEMI~\cite{wei2019semi}                     & 22.35          & 0.788          & 16.56          & 0.486          & 25.03          & 0.842          & 24.43          & 0.782          & 26.05          & 0.822          & 22.88          & 0.744          \\
DIDMDN~\cite{zhang2018density}                 & 22.56          & 0.818          & 17.35          & 0.524          & 25.23          & 0.741          & 28.13          & 0.867          & 29.65          & 0.901          & 24.58          & 0.770          \\
UMRL~\cite{yasarla2019uncertainty}                    & 24.41          & 0.829          & 26.01          & 0.832          & 29.18          & 0.923          & 29.97          & 0.905          & 30.55          & 0.910          & 28.02          & 0.880          \\
RESCAN~\cite{li2018recurrent}               & 25.00          & 0.835          & 26.36          & 0.786          & 29.80          & 0.881          & 31.29          & 0.904          & 30.51          & 0.882          & 28.59  &         0.857          \\
PreNet~\cite{ren2019progressive}                    & 24.81          & 0.851          & 26.77          & 0.858          & 32.44          & 0.950          & 31.75          & 0.916          & 31.36          & 0.911          & 29.42          & 0.897          \\
MSPFN~\cite{jiang2020multi}                     & 27.50          & 0.876          & 28.66          & 0.860          & 32.40          & 0.933          & 32.82          & 0.930          & 32.39          & 0.916          & 30.75          & 0.903          \\
MPRNet~\cite{zamir2021multi}                    & 30.27          & 0.897          & 30.41          & 0.890          & 36.40          & 0.965          & 33.64          & 0.938          & 32.91          & 0.916          & 32.73          & 0.921          \\
SPAIR~\cite{purohit2021spatially}                     & 30.35          & 0.909          & 30.95          & 0.892          & 36.93          & 0.969          & 33.34          & 0.936          & 33.04          & \underline{0.922}          & 32.91          & 0.926          \\
Restormer~\cite{zamir2022restormer}                 & \textbf{32.00}          & \textbf{0.923}          & \underline{31.46}          & \underline{0.904}          & \underline{38.99}          & \underline{0.978}          & \underline{34.18}          & \underline{0.944}          & \underline{33.19}          & \textbf{0.926}          & \underline{33.96}          & \textbf{0.935}          \\ \hline
Ours-L V1~\cite{qiu2023mb}  & 31.48        & \underline{0.917}         &  31.28      & 0.903           &    38.60          &\textbf{0.980}             & 34.00             & 0.942  & 32.93            & 0.917  & 33.66            & \underline{0.932}     \\ 
Ours-L V2  & \underline{31.88}          & \textbf{0.923}          & \textbf{31.57}          & \textbf{0.909}          & \textbf{39.03}         & \textbf{0.980}          & \textbf{34.20}             & \textbf{0.946}             &    \textbf{33.31}           &0.919              & \textbf{34.00}              & \textbf{0.935}              \\ \bottomrule
\end{tabular}}
\end{table*}
\begin{figure*}[t]
    \centering
    \includegraphics[width=18cm]{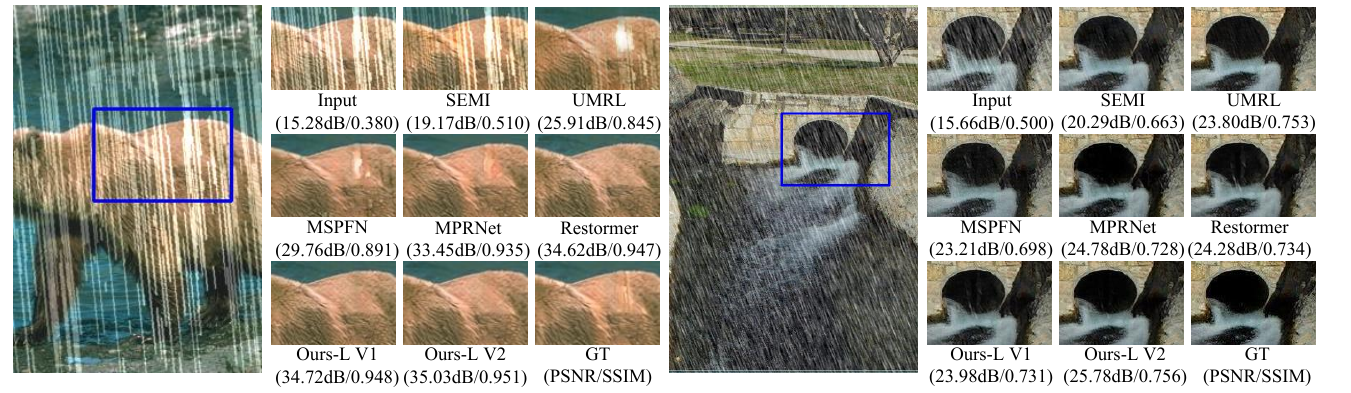}
  \caption{\textbf{Image deraining on the images ”1” and ”282” from Rain100H~\cite{yang2017deep} and Test1200~\cite{zhang2018density}.} Our MB-TaylorFormer-L V2 generates rain-free images with structural fidelity and without artifacts. }
    \label{fig:derain}
    \vspace{-0.2cm}
\end{figure*}

\begin{table*}[]
    \centering
\caption{\textbf{Quantitative comparisons of benchmark models on desnowing datasets.}``-'' indicates that the result is not available. The best and second-best results are highlighted in bold and underlined, respectively.}
\label{tab:desnow}
\scalebox{0.9}{
\centering
\begin{tabular}{cc|cccccccc|cc}
\toprule
\multicolumn{2}{c|}{Models}                             & \begin{tabular}[c]{@{}c@{}}DsnowNet\\~\cite{liu2018desnownet}\end{tabular}  & \begin{tabular}[c]{@{}c@{}}JSTASR\\~\cite{chen2020jstasr}\end{tabular}  &\begin{tabular}[c]{@{}c@{}}HDCW-Net\\~\cite{chen2021all}\end{tabular}   & 
\begin{tabular}[c]{@{}c@{}}SMGARN\\~\cite{cheng2023snow}\end{tabular} &
\begin{tabular}[c]{@{}c@{}}MSP-Former\\~\cite{chen2023msp}\end{tabular} &
\begin{tabular}[c]{@{}c@{}}FocalNet\\~\cite{cui2023focal}\end{tabular} &
\begin{tabular}[c]{@{}c@{}}IRNeXt\\~\cite{cui2023irnext}\end{tabular} &
\begin{tabular}[c]{@{}c@{}}ConvIR-B\\~\cite{cui2024revitalizing}\end{tabular} &
\begin{tabular}[c]{@{}c@{}}Ours-L V1\\~\cite{qiu2023mb}\end{tabular} & Ours-L V2      \\ \hline
\multicolumn{1}{c|}{\multirow{2}{*}{Snow100K}}     & PSNR$\uparrow$    & 30.50    & 23.12  & 31.54                          & 31.92  & 33.43        & 33.53  & 33.61  & \underline{33.92}       & 33.79                     & \textbf{34.01} \\
\multicolumn{1}{c|}{}                          & SSIM$\uparrow$    & 0.94     & 0.86   & 0.95                          & 0.93   & \textbf{0.96}       & \underline{0.95}     & \underline{0.95}   & \textbf{0.96}      & \underline{0.95}                          & \textbf{0.96}  \\ \hline
\multicolumn{1}{c|}{\multirow{2}{*}{SRRS}}      & PSNR$\uparrow$    & 20.38   & 25.82  & 27.78                        & 29.14  & 30.76        & 31.34  & 31.91  & \underline{32.39}       & 32.26                     & \textbf{32.55} \\
\multicolumn{1}{c|}{}                          & SSIM$\uparrow$    & 0.84    & 0.89   & 0.92                        & 0.94   & \underline{0.95}       & \textbf{0.98}    & \textbf{0.98}   & \textbf{0.98}      & \textbf{0.98}                         & \textbf{0.98}  \\ \hline
\multicolumn{1}{c|}{\multirow{2}{*}{overhead}} & \#Param$\downarrow$ & 15.6M    & 65M    & 6.99M     & 6.83M  & \textbf{2.83M}     & \underline{3.74M}    & 5.46M  & 8.63M         & 7.43M                           & 7.29M          \\
\multicolumn{1}{c|}{}                          & MACs$\downarrow$    & \textbf{1.7K}     & -      & 9.8G     & 450.3G & \underline{4.4G}     & 30.6G   & 42.1G & 71.2G          & 88.1G                        & 86.0G         \\ \bottomrule
\end{tabular}}
\end{table*}

\begin{figure*}[t]
    \centering
    \includegraphics[width=18cm]{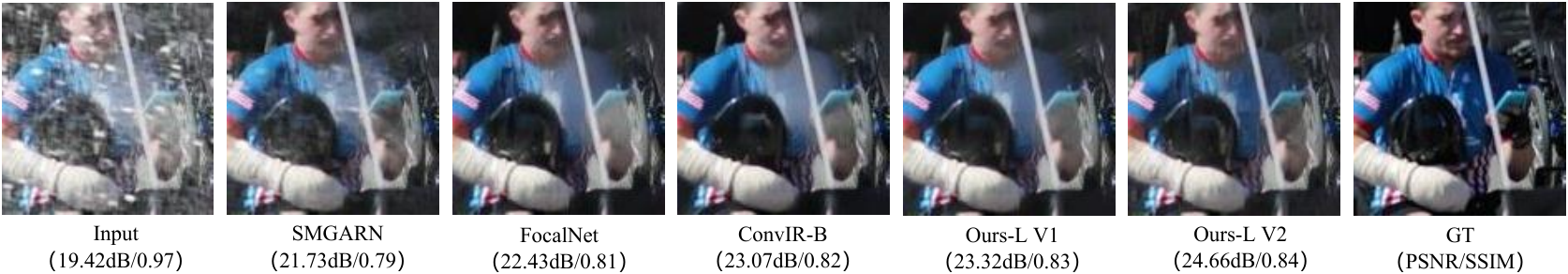}
  \caption{\textbf{Image desnowing on the images "city\_read\_05583" from Snow100K~\cite{liu2018desnownet}.} Our MB-TaylorFormer-L V2 generates cleaner snow-free images. ``Ours V1" represents the conference version of our MB-TaylorFormer~\cite{qiu2023mb}.}
    \label{fig:desnow}
    \vspace{-0.3cm}
\end{figure*}

% \usepackage{multirow}

\begin{figure*}[t]
    \centering
    \includegraphics[width=18cm]{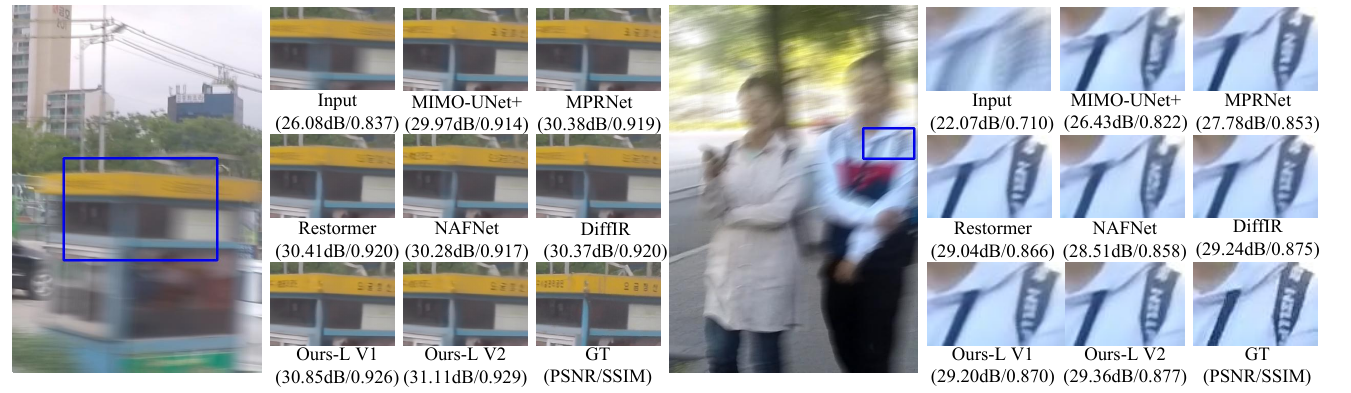}
  \caption{\textbf{Image motion deblurring comparison on the images "GOPR0410\_11\_00-000190" and "55fromGOPR1096.MP4" from GoPro~\cite{nah2017deep} and HIDE~\cite{shen2019human}.} Our MB-TaylorFormer-XL V2 obtains sharper and visually-faithful results.}
  \label{fig:deblur_comparision}
    \label{fig:MOTION}
    \vspace{-0.2cm}
\end{figure*}

\subsection{Image Dehazing Results}

We progressively train our MB-TaylorFormer V2 with the same settings as ~\cite{qiu2023mb} on synthetic datasets (ITS~\cite{8451944}, OTS~\cite{8451944} and HAZE4K~\cite{liu2021synthetic}) and real-world datasets (O-HAZE~\cite{ancuti2018haze}, Dense-Haze~\cite{ancuti2019dense}, and NH-HAZE~\cite{ancuti2021ntire}). The quantitative results in Tab.~\ref{tab:dehze} and Fig.~\ref{fig:comparison_four}(a) highlight that our model significantly outperforms other models. Specifically, our MB-TaylorFormer-L V2 achieves a 0.12dB and 0.77dB improvement in PSNR over the recent SOTA model ConIR-B~\cite{cui2024revitalizing} on the synthetic datasets ITS and Haze4K, respectively, while utilizing only 84.5\% of the number of parameters of ConIR-B. On the outdoor synthetic dehazing dataset OTS, our MB-TaylorFormer-L V2 achieves the second-best performance, significantly outperforming the subsequent methods C$^{2}$PNet~\cite{zheng2023curricular} and ConvIR-S~\cite{cui2024revitalizing}. Furthermore, for small-scale real-world datasets O-HAZE and NH-Haze, our MB-TaylorFormer-L V2 achieves PSNR/SSIM gains of 0.07dB/0.012 and 0.11dB/0.014 over the previous best-performing model ConvIR, respectively. This indicates that MB-TaylorFormer V2 has strong capability and generalization for image dehazing. Compared to MB-TaylorFormer V1~\cite{qiu2023mb}, MB-TaylorFormer V2 with the same scale achieves improved performance, indicating that the proposed T-MSA++ is effective and focuses more on crucial regions. We also show the visual results of MB-TaylorFormer-L V2 in comparison to other SOTA dehazing models. As depicted in Fig.~\ref{fig:haze} and Fig.~\ref{fig:dehaze2}, the comparison highlights a stark difference between the shadows in images produced by counterpart models and those from our approach. Notably, the counterparts suffer from evident artifacts and texture degradation, resulting in less natural shadows. Conversely, our method yields dehazed images characterized by heightened clarity, enhanced cleanliness, and a remarkable resemblance to the ground truth.

\subsection{Image Deraining Results}
Following previous work~\cite{zamir2022restormer}, we train our model on 13,712 clean-rain image pairs collected from multiple datasets~\cite{ren2019progressive,wei2019semi,yasarla2019uncertainty,zhang2018density,li2018recurrent,fu2017clearing}. With this single trained model, we conduct evaluations on various test sets, including Rain100H~\cite{yang2017deep}, Rain100L~\cite{yang2017deep}, Test100~\cite{zhang2019image}, Test2800~\cite{fu2017clearing}, and Test1200~\cite{zhang2018density}. We calculate PSNR(dB)/SSIM scores using the Y channel within the YCbCr color space. The results in Tab.~\ref{tab:derain} highlight the consistent and comparable performance improvements achieved by our MB-TaylorFormer-L V2 across all five datasets. In comparison to the recent SOTA model Restormer~\cite{zamir2022restormer}, MB-TaylorFormre-L V2 achieves optimal or suboptimal performance across all datasets. On specific datasets, such as Test1200~\cite{zhang2018density}, the improvement can be as significant as 0.12dB, while utilizing
only $62.5\%$ of the MACs compared to Restormer. In addition, compared to MB-TaylorFormer-L V1~\cite{qiu2023mb}, the average PSNR has increased by 0.34dB, indicating the effectiveness of the proposed T-MSA++. The challenging visual examples are presented in Fig.~\ref{fig:derain} and Fig.~\ref{fig:comparison_four}(b), where MB-TaylorFormer-L V2 can generate raindrop-free images while preserving the underlying structural content.

\subsection{Image Desnowing Results}
We conduct desnowing experiments on a synthesis dataset.
Specifically, we train and test on the Snow100K~\cite{liu2018desnownet} and SRRS~\cite{chen2020jstasr} dataset, respectively.
We compare the performance of MB-TaylorFormer-L V2 with other methods and report the results in Tab.~\ref{tab:desnow}. On Snow100K dataset, MB-TaylorFormer-L V2 outperforms the previous best method, ConvIR-B, by 0.09dB and exceeds the IRNeXt method by 0.4dB in terms of PSNR. On SRRS dataset, MB-TaylorFormer-L V2 outperforms the previous best method, ConvIR-B, by 0.16dB in terms of PSNR. Additionally, the number of parameters of our model is only 84.5\% of that of ConvIR-B. Fig.~\ref{fig:desnow} presents a visual comparison. Our approach effectively avoids artifacts and performs better in removing snow. This is attributed to the larger receptive field of the Transformer, which allows for long-range interactions to gather information from distant areas for image restoration. In contrast, convolutional methods have difficulties in handling degradation over large areas.

% \usepackage{multirow}

\begin{table}[t]
\caption{\textbf{Quantitative comparisons of benchmark models on debluring datasets.} ``-" indicates that the result is not available. The best and second-best results are highlighted in bold and underlined, respectively.}
\label{tab:deblur}
\centering
\begin{tabular}{c|cc|cc}
\toprule
\multirow{2}{*}{Models} & \multicolumn{2}{c|}{GoPro~\cite{nah2017deep}}      & \multicolumn{2}{c}{HIDE~\cite{shen2019human}}      \\ \cline{2-5} 
                         & PSNR$\uparrow$ & SSIM$\uparrow$ & PSNR$\uparrow$ & SSIM$\uparrow$  \\ \hline
DeblurGAN-v2~\cite{kupyn2019deblurgan}             & 29.55          & 0.934          & 26.61          & 0.875            \\
SRN~\cite{tao2018scale}                      & 30.26          & 0.934          & 28.36          & 0.915              \\
DBGAN~\cite{zhang2020deblurring}                    & 31.10          & 0.942          & 28.94          & 0.915                  \\
MT-RNN~\cite{park2020multi}                   & 31.15          & 0.945          & 29.15          & 0.918              \\
DMPHN~\cite{zhang2019deep}                    & 31.20          & 0.940          & 29.09          & 0.924                 \\
SPAIR~\cite{purohit2021spatially}                     & 32.06          & 0.953          & 30.29          & 0.931                  \\
MIMO-UNet+~\cite{cho2021rethinking}               & 32.45          & 0.957          & 29.99          & 0.930             \\
IPT~\cite{chen2021pre}                      & 32.52          & -              & -              & -                   \\
MPRNet~\cite{zamir2021multi}                    & 32.66          & 0.959          & 30.96          & 0.939              \\
Restormer~\cite{zamir2022restormer}                & 32.92          & \underline{0.961}          & 31.22          & 0.942         \\
NAFNet~\cite{chen2022simple}                  & 33.03          & \underline{0.961}          & 31.32          & 0.941           \\
DiffIR~\cite{xia2023diffir}                 & 33.20          & \textbf{0.963}          & \underline{31.55}          & \textbf{0.947}                     \\
ConIR-L~\cite{cui2024revitalizing}                 & \textbf{33.28}         & \textbf{0.963}         & 30.92         & 0.937                     \\ \hline
Ours-XL V1~\cite{qiu2023mb}  & 32.95      & 0.960          & 31.33         & 0.942 \\
Ours-XL V2  & \underline{33.24}       & \textbf{0.963}          & \textbf{31.66}          & \underline{0.946}               \\ \bottomrule
\end{tabular}
    \vspace{-0.3cm}
\end{table}

\begin{figure*}[t]
    \centering
    \includegraphics[width=18cm]{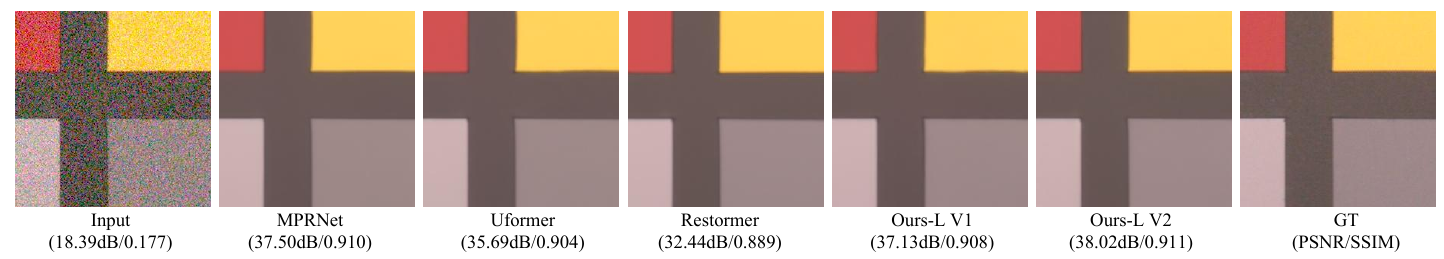}
  \caption{\textbf{Image denoising on the image "0003-0030" from SIDD dataset~\cite{abdelhamed2018high}.} Our MB-TaylorFormer-L V2 generates noise-free images with structural fidelity and without artifacts. }
    \label{fig:denoise_compare}
    \vspace{-0.2cm}
\end{figure*}

\begin{table*}[ht]
    \centering
\caption{\textbf{Quantitative comparisons of benchmark models on denoising datasets.} ``-" indicates that the result is not available. The best and second-best results are highlighted in bold and underlined, respectively.}
\scalebox{0.85}{
\centering
\begin{tabular}{cc|cccccccccc|cc}
\toprule
\multicolumn{2}{c|}{Models}                     & \begin{tabular}[c]{@{}c@{}}VDN\\~\cite{yue2019variational}\end{tabular}   & \begin{tabular}[c]{@{}c@{}}SADNet\\~\cite{chang2020spatial}\end{tabular} & \begin{tabular}[c]{@{}c@{}}DANet+\\~\cite{yue2020dual}\end{tabular} & \begin{tabular}[c]{@{}c@{}}CycleISP\\~\cite{zamir2020cycleisp}\end{tabular} & \begin{tabular}[c]{@{}c@{}}MIRNet\\~\cite{zamir2020learning}\end{tabular} & \begin{tabular}[c]{@{}c@{}}DeamNet\\~\cite{ren2021adaptive}\end{tabular} & \begin{tabular}[c]{@{}c@{}}MPRNet\\~\cite{zamir2021multi}\end{tabular} & \begin{tabular}[c]{@{}c@{}}DAGL\\~\cite{mou2021dynamic}\end{tabular}  & \begin{tabular}[c]{@{}c@{}}Uformer\\~\cite{wang2022uformer}\end{tabular} & \begin{tabular}[c]{@{}c@{}}Restormer\\~\cite{zamir2022restormer}\end{tabular}  & \begin{tabular}[c]{@{}c@{}}Ours-L V1\\~\cite{qiu2023mb}\end{tabular}  & Ours-L V2 \\ \hline
\multicolumn{1}{c|}{\multirow{2}{*}{SIDD}} & PSNR$\uparrow$    & 39.28 & 39.46  & 39.47  & 39.52    & 39.72  & 39.47   & 39.71  & 38.94 & 39.77   & \underline{40.02}   & 39.98   & \textbf{40.11}     \\
\multicolumn{1}{c|}{}                      & SSIM$\uparrow$    & 0.956 & 0.957  & 0.957  & 0.957    & \underline{0.959}  & 0.957   & 0.958  & 0.953 & \underline{0.959}   & \textbf{0.960}  & \underline{0.959}     & \textbf{0.960}      \\ \hline
\multicolumn{1}{c|}{\multirow{2}{*}{overhead}} & Params$\downarrow$                                                          & 7.81M                                                 & 4.32M                                                    & 9.15M                                                    & \underline{2.83M}                                                & 31.8M                                              & \textbf{2.22M}                                               & 11.3M                                              & 5.62M                                            & 20.63M                                                & 26.10M                                                 & 7.43M                                                                      & 7.29M                \\
\multicolumn{1}{c|}{}                          & MACs$\downarrow$                                                       & 49.4G                                                 & \underline{21.4G}                                                    & \textbf{14.8G}                                                    & 335.0G                                               & 785.0G                                             & 145.8G                                              & 571.2G                                             & 22.8G                                            & 43.9G                                                & 141.0G                                                 & 88.1G                                                                      & 86.0G                \\ \bottomrule
\end{tabular}}
\label{denoise}
\end{table*}

\subsection{Image Motion Deblurring Results}

MB-TaylorFormer-XL V2 is trained on the GoPro dataset~\cite{nah2017deep} for the task of image motion deblurring. Subsequently, the performance of MB-TaylorFormer-XL V2 is assessed on two established datasets: GoPro and HIDE~\cite{shen2019human}. We compare MB-TaylorFormer-XL V2 with the SOTA image motion deblurring models, including Restormer~\cite{zamir2022restormer}, NAFNet~\cite{chen2022simple}, and DiffIR~\cite{xia2023diffir}. The quantitative results, encompassing PSNR and SSIM metrics, are presented in Tab.~\ref{tab:deblur}. Notably, our MB-TaylorFormer-XL V2 exhibits superior performance compared to other motion deblurring models. Specifically, on the GoPro dataset, which is regarded as a difficult dataset, MB-TaylorFormer-XL V2 outperforms Restormer~\cite{zamir2022restormer} and NAFNet~\cite{chen2022simple} by 0.32dB and 0.21dB, respectively. Furthermore, when compared to DiffIR on both GoPro and HIDE datasets, MB-TaylorFormer-XL V2 demonstrates improvements by 0.04dB and 0.11dB. These results underscore the efficacy of MB-TaylorFormer V2 in achieving SOTA motion deblurring performance. 
Furthermore, while MB-TaylorFormer-XL V1 achieves competitive PSNR values of 32.95dB on GoPro and 31.33dB on HIDE, our MB-TaylorFormer-XL V2 demonstrates superior performance, surpassing MB-TaylorFormer-XL V1~\cite{qiu2023mb} by 0.29dB on GoPro and 0.33dB on HIDE, respectively. The qualitative results are shown in Fig.~\ref{fig:deblur_comparision} and Fig.~\ref{fig:comparison_four}(c). It is worth noting that MB-TaylorFormer-XL V2 exhibits the highest visual quality, especially in restoring tiny text details with enhanced clarity. This qualitative assessment aligns with the quantitative research findings, further confirming the exceptional performance of our MB-TaylorFormer-XL V2.

% \usepackage{multirow}

\subsection{Image Denoising Results}

We conduct denoising experiments on a real-world dataset. Specifically, we train and test on the SIDD dataset~\cite{abdelhamed2018high}. Consistent with prior studies~\cite{zamir2022restormer}, denoising is executed using the bias-free MB-TaylorFormer-L V2 model, which allows for adaptation to a broad range of noise levels. Tab.~\ref{denoise} shows the superior performance of our model. Particularly, on the SIDD dataset, our MB-TaylorFormer-L V2 achieves noteworthy PSNR gains, surpassing the previous leading CNN model MPRNet~\cite{chen2022simple} by 0.4dB. 
Additionally, compared to MB-TaylorFormer-L V1~\cite{qiu2023mb}, the PSNR gain of MB-TaylorFormer-L V2 reaches as high as 0.13dB. The visual results depicted in Fig.~\ref{fig:denoise_compare} show that our MB-TaylorFormer-L V2 excels in producing clean images while preserving fine textures, indicating that compared to CNN methods, MB-TaylorFormer-L V2 can effectively utilize the low-pass characteristics of Transformers to filter out high-frequency noise. Additionally, compared to some Transformer methods, MB-TaylorFormer-L V2 can leverage its more precise global modeling capability to achieve better results. Fig.~\ref{fig:comparison_four}(d) demonstrates our advantage in computational cost.

% \usepackage{multirow}
% \usepackage{multirow}

% \usepackage{multirow}

\begin{figure*}[t]
  \centering
  \captionsetup[subfloat]{labelformat=empty}
    \subfloat[(a) Conv 
  ]{
    \includegraphics[height=1.25in]{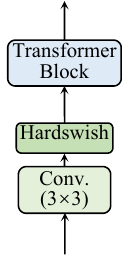}
  }
  \hspace{-15mm}
  \subfloat[\qquad \quad (b) Conv-P
  ]{
    \hspace{0.5in}
    \includegraphics[height=1.25in]{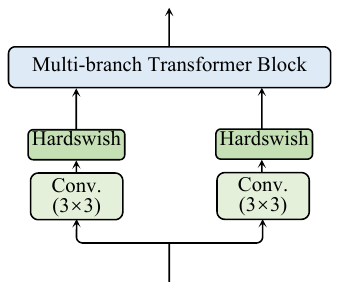}
  }
  \hspace{-15mm}
% kind++  27.09/0.9469
  \subfloat[\qquad \quad (c) Dilated Conv-P
  ]{
    \hspace{0.5in}
    \includegraphics[height=1.25in]{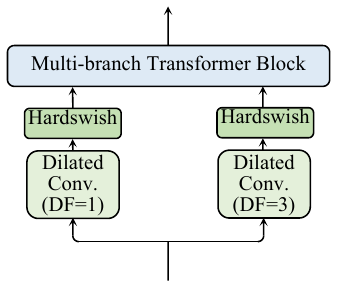}
  }
  \hspace{-15mm}
    \subfloat[\qquad \quad (d) Conv-SP
  ]{
    \hspace{0.5in}
    \includegraphics[height=1.25in]{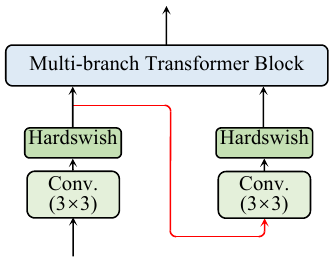}
  }
  \hspace{-15mm}
  \subfloat [\qquad \quad (e) DSDCN-SP
  ]{
   \hspace{0.5in}
    \includegraphics[height=1.25in]{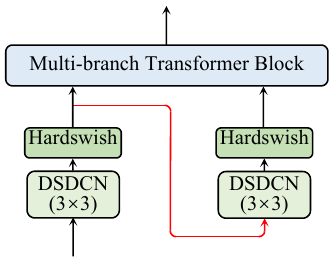}
  }
\caption{The structure of patch embedding.}
        \label{fig:fig1_1}
            \vspace{-0.2cm}
\end{figure*}

\begin{table}[t]
\centering
\caption{\textbf{Ablation studies for the multi-scale patch embedding and multi-branch structure.} ``-SP" means two convolutional layers simultaneously in series and parallel with the same kernel size of 3, and ``-P" means two convolutional layers in parallel with the same kernel size of 3.}
\label{tab:tab2}
\scalebox{1}{
\begin{tabular}{c|c|cc|c|c}
\toprule
Branch & Type of Conv. & PSNR                            & SSIM                            & Params & MACs   \\ \hline
Single & Conv         & 39.43                           & 0.992                           & 2.61M   & 32.8G \\ \hline
\multirow{4}{*}{Double} & Conv-P         & 39.87                          & 0.991                           & 2.60M   & 37.1G \\
& Dilated Conv-P & 39.99                          & 0.992                           & 2.60M   & 37.1G \\
      & Conv-SP         & 40.25                           & 0.992                           & 2.60M   & 37.1G \\
       & DSDCN-SP         & \textbf{41.00} & \textbf{0.993} & 2.63M   & 37.7G  \\ \bottomrule
\end{tabular}}
%\vspace{-0.1cm}
\end{table}

 \begin{table}[t]
 \centering
\caption{\textbf{Ablation experiments for the self-attention.} We compare T-MSA++ with other linear self-attention modules and investigate the effect of CPE.}
\label{tab:tab5}
\scalebox{0.92}{
\begin{tabular}{c|cc|c|c}
\toprule
Models                     & PSNR  & SSIM  & Params & MACs   \\ \hline
MB-TaylorFormer-B V2 & \textbf{41.00} & \textbf{0.993} & 2.63M & 37.7G \\\hline
(a) w/o CPE   & 39.76 & 0.991 & 2.58M & 37.6G\\
(b) CPE $\to$ MSAR~\cite{qiu2023mb}   & 40.73 & 0.992 & 2.69M & 38.5G   \\ \hline
(c) T-MSA++ $\to$ MDTA~\cite{zamir2022restormer}                  & 38.57 & 0.991 & 2.58M & 35.2G \\
(d) T-MSA++ $\to$ Swin~\cite{liang2021swinir}                  & 36.59 & 0.988 & 2.52M & 36.4G \\ 
(e) T-MSA++ $\to$ TAR~\cite{katharopoulos2020transformers}                 & 36.74 & 0.987 & 2.52M & 34.0G \\ 
(f) T-MSA++ $\to$ PVTv2~\cite{wang2022pvt}                 & 38.10 & 0.990 & 10.89M & 38.6G \\ 
(g) T-MSA++ $\to$ Cswin~\cite{dong2022cswin}                 & 38.19 & 0.987 & 3.30M & 40.1G \\ 
(h) T-MSA++ $\to$ LinFormer~\cite{wang2020linformer}                 & 36.12 & 0.983 & 48.70M & 371.7G \\ 
(i) T-MSA++ $\to$ Flatten~\cite{han2023flatten}                 & 40.47 & 0.993 & 2.63M & 36.7G \\
\bottomrule
\end{tabular}}
\vspace{-0.1cm}
\end{table}

\begin{table}[htb]
 \centering
\caption{\textbf{Study of the remainder.} The smaller approximation error for Softmax-attention of Swin~\cite{liu2021swin}, the better the performance.}
\label{tab:tab7}
\scalebox{1.1}{
\begin{tabular}{c|cc}
\toprule
Models                      & PSNR  & SSIM    \\ \hline
Swin~\cite{liu2021swin} & \textbf{36.59} & \textbf{0.988}  \\
Swin + T-MSA++(2nd)~\cite{qiu2023mb}                     & 36.50 & \textbf{0.988}   \\
Swin + T-MSA++(1st)~\cite{qiu2023mb}                      &          36.37  &   0.987     \\ \bottomrule
\end{tabular}}
\vspace{-0.1cm}
\end{table}

\begin{table}[htb]
\centering
\caption{\textbf{Analysis of the truncation range of offsets.} Local correlations of tokens can improve model performance.}
\label{tab:tab-0}
\scalebox{1}{
\begin{tabular}{c|cc}
\toprule
Truncation range & PSNR  & SSIM  \\ \hline
w/o              & 40.13 & 0.991 \\ 
{[}-2, 2{]}       & 40.88 & 0.992 \\ 
{[}-3, 3{]}       & \textbf{41.00} & \textbf{0.993} \\ 
{[}-4, 4{]}       & 40.53 & 0.992 \\ \bottomrule
\end{tabular}
}
\vspace{-0.2cm}
\end{table}

\subsection{Ablation Study}

In this section, we conduct ablation experiments on the MB-TaylorFormer-B V2 model using the dehazing ITS~\cite{8451944} dataset to assess and understand the robustness and effectiveness of each module of the model.

\textbf{Exploration of multi-scale patch embedding and multi-branch structures.} In Tab.~\ref{tab:tab2}, we investigate variations in patch embedding and the impact of employing different numbers of branches. Our baseline model is a single-branch configuration based on standard single-scale convolution, as shown in Fig.~\ref{fig:fig1_1}(a). We then introduce modifications in the following ways. 1) \textit{Multi-Branch Structure:} To assess the influence of a multi-branch structure, we design patch embedding models with a single-scale convolution and multi-branch parallel configuration (Conv-P), as shown in Fig.~\ref{fig:fig1_1}(b). 2) \textit{Multiple Receptive Field Sizes:} To explore the effect of multiple receptive field sizes, we incorporate parallel dilated convolutional layers (DF=1, 2) for patch embedding (Dilated Conv-P), as shown in Fig.~\ref{fig:fig1_1}(c). 3) \textit{Multi-Level Semantic Information Investigation:} To delve into the impact of multi-level semantic information, we replace dilated convolution with standard convolution for patch embedding, employing a series connection between two convolutional layers (Conv-SP), as shown in Fig.~\ref{fig:fig1_1}(d). 4) \textit{Flexible Receptive Field Shape Examination:} To assess the impact of flexible receptive field shapes, we substitute standard convolution with DSDCN (DSDCN-SP), as shown in Fig.~\ref{fig:fig1_1}(e). The experimental results indicate that the performance, ranked from the best to the worst, follows the order: DSDCN-SP, Conv-SP, Dilated Conv-P, Conv-P, and Conv. This suggests that our multi-scale patch embedding approach offers flexibility in patch representation.

% \usepackage{multirow}
% \usepackage{multirow}
% \usepackage{multirow}

%\noindent
\textbf{Effectiveness of convolutional positional encoding}. Tab.~\ref{tab:tab5}(a) demonstrates our convolutional positional encoding (CPE) module provides a favorable gain of $1.24$dB over the counterpart without CPE module, with only a tiny increase in the number of parameters ($0.05$M) and MACs ($0.45$G), which is attributed to the fact that the CPE module provides a higher rank of the attention map~\cite{han2023flatten} as well as the relative positional information of the tokens. In addition, Tab.~\ref{tab:tab5}(b) indicates that compared to our previously adopted MSAR module~\cite{qiu2023mb} used for providing local error correction and position encoding, CPE achieves a gain of 0.27dB in PSNR. This indicates that the T-MSA++ can efficiently approximate the higher-order cosine terms without the help of MSAR and that the CPE module is more suitable for T-MSA++.

%\\
\textbf{Comparison with other linear self-attention modules}. 
Tab.~\ref{tab:tab5}(c)-(i) presents a comparison between the proposed T-MSA++ and several common linear self-attention modules. The results indicate that TaylorFormer exhibits significant advantages over existing linear self-attention modules. This is attributed to several reasons: 1) T-MSA++ has a finer-grained self-attention capability compared to MDTA~\cite{zamir2022restormer}; 2) our model excels at modeling long-distance pixels compared to Cswin~\cite{dong2022cswin} and Swin~\cite{liu2021swin}; 3) the pooling mechanism in PVTv2~\cite{wang2022pvt} leads to information loss; 4) LinFormer~\cite{wang2020linformer} relies on constructing a learnable low-rank matrix, which results in a high parameter count while limiting the rank of the attention map.

\begin{figure}[htb]
    \centering
    \includegraphics[width=9cm]{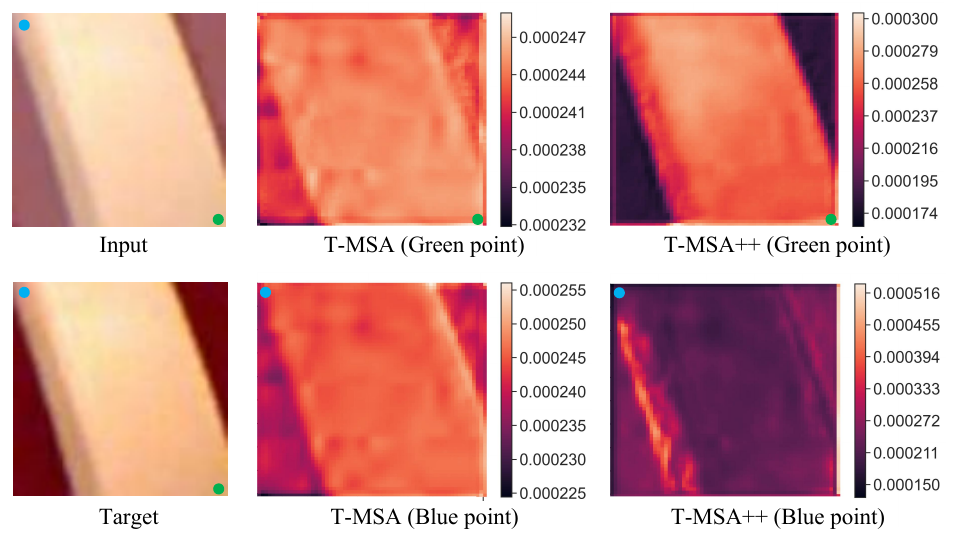}
    \caption{\textbf{Attention maps from T-MSA and T-MSA++ for an image patch.} The attention map is computed based on the queries corresponding to the blue and green points, and the keys corresponding to all points. Through the distribution of attention and colorbar, it is evident that T-MSA++ is capable of generating sharp attention, while attention of T-MSA is relatively smooth.}
    \label{fig:figure10}
    \vspace{-0.3cm}
\end{figure}

%\\
\textbf{Analysis of the remainder of the Taylor expansion.} To explore the reminder and its impact, we investigate the effect of different orders of Taylor expansion for Softmax-attention. Considering that the associative law is not applicable to the second-order Taylor expansion of T-MSA++ or T-MSA~\cite{qiu2023mb} (T-MSA-2nd), which leads to a significant computational burden, we perform first-order and second-order Taylor expansions for Swin. Tab.~\ref{tab:tab7} shows that T-MSA-1st can approximate the performance of Softmax-attention, while higher-order Taylor expansions, such as T-MSA-2nd, can better approximate Softmax-attention. However, the computational complexity of T-MSA-2nd increases quadratically with image resolution, so it is difficult to model long-range dependence in practical applications. Unlike the quadratic computational complexity of T-MSA-2nd, T-MSA++ is an algorithm with linear computational complexity. Fig.~\ref{fig:figure10} visualizes the attention map of the first layer of the model. We observe that for the point on the front of the chair (green point), the points corresponding to the front of the chair in the attention map have higher weights, while for the points on the side of the chair (blue point), the corresponding points in the attention map have higher weights. This indicates that T-MSA++ can focus on more crucial regions. This indicates that T-MSA++ has acquired the ability of attention focus similar to Softmax-attention.

\textbf{The truncation range of offsets.} Tab.~\ref{tab:tab-0} shows the effect of different truncation ranges on the model. We find DSDCN with truncated offsets achieves better performance than DSDCN without truncated offsets. We attribute the improvement to the fact that the generated tokens in our approach focus more on local areas of the feature map. We further investigate the effect of different truncation ranges and finally choose $[\text{-}3, 3]$ as the truncation range for MB-TaylorFormer V2.

\begin{figure}[ht]
    \centering
    \includegraphics[width=9cm]{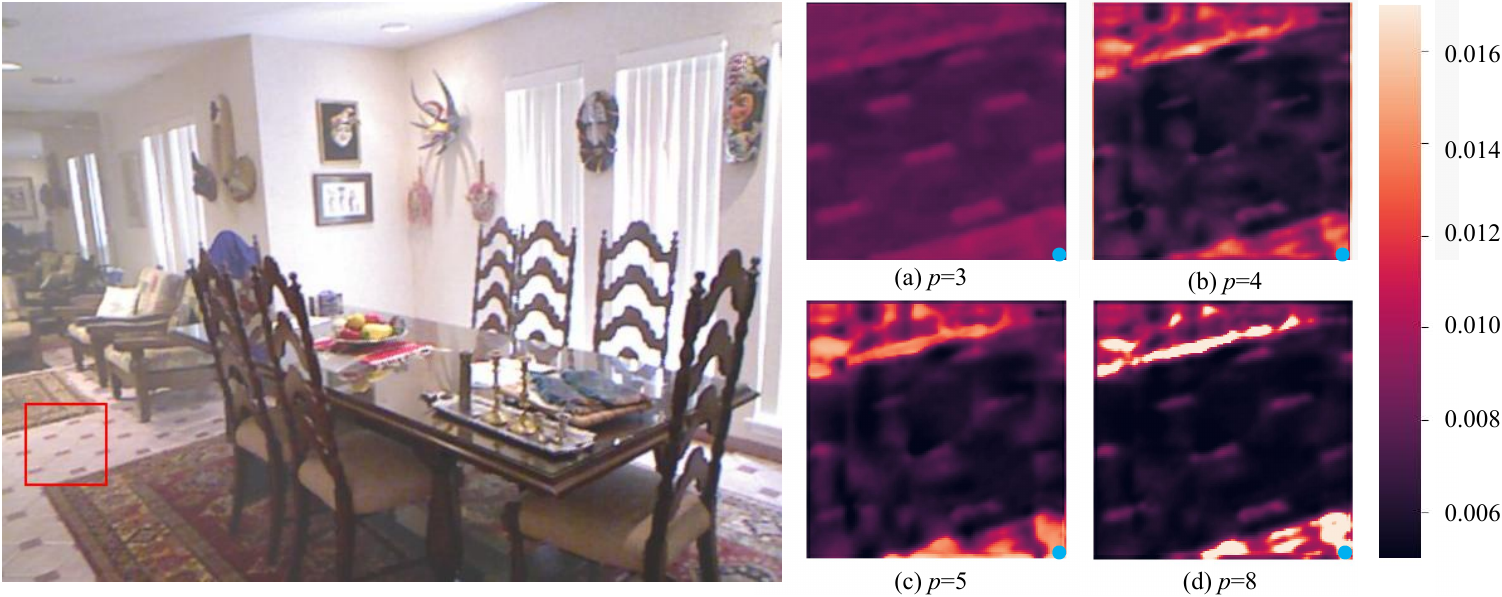}
    \caption{\textbf{ Visualization of attention maps in the first layer of the network with varying '$p$' values.} A larger $p$ makes the model pay more attention to regions that are similar to the blue points.}
    \label{fig:figurepppp}
    \vspace{-0.2cm}
\end{figure}

\textbf{The choose of focused factor '$p$'.} The performance of our model is robust to variations in '$p$'. Specifically, when '$p$' ranges from 3 to 8, the PSNR/SSIM does not change significantly (refer to Tab.~\ref{tab:tabppppp}). This indicates that the model is not sensitive to this hyper-parameter. The attention maps of the network first layer are shown in  Fig.~\ref{fig:figurepppp} and as '$p$' increases, the attention of the model to key areas is significantly enhanced. This allows the model to accurately capture important features in the images, thereby improving overall performance. However, when '$p$' exceeds a certain threshold, the enhancement effect of this attention tends to plateau and may cause over-focusing on local details while neglecting global information. Referring to Fig.~\ref{fig:figurepppp}, $p$=4 is a balanced choice, as it allows for smooth focusing on important regions. Therefore, for simplicity, we select $p$=4 for all models presented in the paper without additional tuning to ensure reliable performance while minimizing the need for extensive hyper-parameter optimization. 

\begin{table}[ht]
    \centering
    \caption{\textbf{Quantitative Comparison of different focused factor $p$.} The model performs best when $p=4$.}
    \label{tab:tabppppp}
    \begin{tabular}{cccccc}
        \toprule
        Focused factor $p$ & 3 & 4 & 5 & 8 \\
        \midrule
        PSNR (dB) & 40.81 & \textbf{41.00} & 40.93 & 40.76 \\
        SSIM & 0.992 & \textbf{0.993} & \textbf{0.993} & 0.992 \\
        \bottomrule
    \end{tabular}
     \vspace{-0.3cm}
\end{table}

\section{Conclusion}
%In this paper, we propose improvements to MB-TaylorFormer, introducing an approximate function for the Taylor expansion error term and incorporating multi-branch parallelism to accelerate inference speed, resulting in the model referred to as MB-TaylorFormer V2. MB-TaylorFormer V2 comprises multi-scale patch embedding and Taylor expansion for Softmax-attention. Multi-scale patch embedding offers a flexible receptive field shape, diverse receptive field sizes, and hierarchical semantic information, enabling the flexible embedding of diverse visual tokens. Additionally, MB-TaylorFormer V2 utilizes the associativity of matrix multiplication for Taylor expansion in Softmax-attention, thereby reducing computational complexity. Furthermore, our mapping function ensures the focusing capability of T-MSA++. Our experiments demonstrate that MB-TaylorFormer V2 is a state-of-the-art image restoration model.%
In previous image restoration methods, there are issues with inaccurate approximations of the original Transformer and the lack of flexibility in tokens. In this work, TaylorFormer V2 is proposed to overcome the shortcomings of existing Linear Transformers, including insufficient receptive field, inability to perform pixel self-attention, and the neglect of value approximations, so it achieves a closer approximation to the original Transformer. In addition, we introduce multi-scale patch embedding to enhance the flexibility of token scales. Additionally, as an improved version of MB-TaylorFormer, we enhance the approximation function for the remainder of the Taylor expansion and adopt a parallel strategy for multiple branches in TaylorFormer V2. This enables MB-TaylorFormer V2 to focus more on crucial areas of the image and improves the inference speed of the model. Experimental results demonstrate that MB-TaylorFormer V2 is a SOTA image restoration model in dehazing, deraining, motion deblurring, and denoising. In the future, we aim to further optimize hardware support for Taylorformer, making our model more hardware-friendly.

\bibliographystyle{IEEEtran}
\bibliography{egbib}
\vspace{-1cm}
\end{document}